\documentclass[letterpaper,9pt]{article}%
\pdfoutput=1
\usepackage{lmodern}
\usepackage{geometry}
\usepackage{authblk}
\usepackage{booktabs}
\usepackage{multirow}
\usepackage[pdftex]{graphicx}
\usepackage{amsmath}
\usepackage{amssymb}

\usepackage{cite}
\usepackage{mathtools}
\usepackage{subfigure}

\newcommand 	\EXP[2] 		{\mathbb{E}_{#1}\hspace{4.0pt}  #2  }
\newcommand 	\EXPC[3]		{\mathbb{E}_{\text{\scriptsize$#1$}\boldsymbol{|}\text{\scriptsize$#2$}}\hspace{4.0pt} #3}
\renewcommand 	\exp[1] 		{\mathrm{exp}\text{$\left( #1 \right)$} }

\newcommand 	\grad[1] 		{\nabla_{\hspace{-1.5pt}#1}}

\newcommand     \MSE[1] 		{\text{MSE}\hspace{1.0pt}(#1)}

\newcommand 	\p[2] 			{p(\hspace{0.8pt} #1 \hspace{1.2pt} \boldsymbol{|} \hspace{1.2pt} #2 \hspace{0.8pt} )}

\newcommand 	\partialfrac[2]	{\frac{\partial #1}{\partial #2}}
\newcommand 	\partialfracsq[2]{\frac{\partial^2 #1}{\partial #2^2}}

\newcommand 	\trace[1] 		{\text{Tr}\hspace{3.0pt}\ #1 }

\newcommand 	\Weibull[2] 	{\mathcal{W}(\hspace{0.8pt} #1 \hspace{1.2pt} \boldsymbol{|} \hspace{1.3pt} #2 \hspace{0.8pt} )}

\def\lim{\mathop{\rm lim}}
\def \kWeibull{k_w}   						%
\def \lWeibull{\lambda_w}  				 	%
\def \PHI {\boldsymbol{\Phi}}					%

\def \Rtheta {R_{\Theta}}					%
\def \x {\mathbf{x}} 						%
\def \y {\mathbf{y}} 						%

\title{Dictionary Learning Strategies for Compressed Fiber Sensing Using a Probabilistic Sparse Model}

\author{Christian Weiss\ }
\author{\ Abdelhak M. Zoubir}

\affil{Signal Processing Group, Institute of Communications and 
       Graduate School of Computational Engineering, Technische Universit\"at Darmstadt, Darmstadt, Germany}%
\date{\small\today}

\begin{document}

\maketitle

\begin{abstract}
\noindent
We present a sparse estimation and dictionary learning framework for compressed fiber sensing based on a probabilistic hierarchical sparse model. 
To handle severe dictionary coherence, selective shrinkage is achieved using a Weibull prior, which can be related to non-convex optimization with
$\ell_p$-norm constraints for $0\!<\!p\!<\!1$. 
In addition, we leverage the specific dictionary structure to promote collective shrinkage based on a local similarity model. %
This is incorporated in form of a kernel function in the joint prior density of the sparse coefficients, thereby establishing a Markov 
random field-relation. %
Approximate inference is accomplished using a hybrid technique that combines Hamilton Monte Carlo and Gibbs sampling. 
To estimate the dictionary parameter, we pursue two strategies, relying on either a deterministic or a 
probabilistic model for the dictionary parameter.
In the first strategy, the parameter is estimated based on alternating estimation.  
In the second strategy, it is jointly estimated along with the sparse coefficients. 
The performance is evaluated in comparison to an existing method in various scenarios using simulations and experimental data. 
\end{abstract}

\section{Introduction}
\noindent
Fiber sensors are versatile devices with broad applicability \cite{Kersey1997,Culshaw2008,Nakazaki2009,Yamashita2009}. 
They are of high interest in smart structures to sense and react to the environment \cite{Measures1992,Udd1996}. %
For quasi-distributed sensing based on wavelength-division \mbox{multiplexing (WDM),} fiber Bragg grating (FBG) sensors are often employed 
due to their sensitivity to strain or \mbox{temperature \cite{Kersey1997,Culshaw2008}.} %
An FBG describes a local variation of the refractive index and reflects light at a certain wavelength, called \emph{Bragg wavelength}. %
Typically, a number of detuned FBGs is imprinted into the core of an optical fiber. Fiber interrogation is performed 
using broadband light sources or wavelength-tunable lasers. 
The latter feature higher local signal-to-noise ratios \mbox{(SNRs) \cite{Nakazaki2009,Yamashita2009}.}
However, in order to monitor time-varying perturbations, the laser has to sweep quickly through the tuning range. %
This requires high-speed analog-to-digital converters (ADCs) %
and produces large amounts of data.\\ %
\emph{Compressed sensing} (CS) \cite{Baraniuk2007,Candes2008,Eldar2012} can help to alleviate these problems by taking samples in form of 
projections into a low-dimensional subspace. 
The original signal can be reconstructed by exploiting the sparsity of the signal with respect to an adequate dictionary \cite{Candes2005,Candes2008a}. %
This task strongly resembles the \emph{sparse synthesis} problem with redundant dictionaries \mbox{in \cite{Tropp2004,Rubinstein2010}.}
Besides greedy methods, such as Orthogonal Matching Pursuit (OMP) \cite{Pati1993}, $\ell_1$-minimization is a popular method to solve 
the sparse reconstruction problem \cite{Tibshirani1994,Donoho2003,Candes2011}. %
It relies on the \emph{restricted isometry property} (RIP), which essentially states that unique sparse solutions can be recovered by restricting 
the $\ell_1$-norm instead of the $\ell_0$-norm \cite{Donoho2001,Candes2006}.
Redundant dictionaries can yield highly sparse representations, that allow for estimating quantities at high resolution 
directly in the sparse domain \cite{Donoho2003,Malioutov2005}. %
However, redundancy causes inter-column coherence and it is likely that the required RIP conditions are no longer fulfilled \cite{Donoho2003,Rauhut2008,Candes2011}.   %
The \mbox{$\ell_p$-norm,} with $0\!<\!p\!<\!1$, offers a trade-off to avoid an NP-hard combinatorical 
problem imposed by the $\ell_0$-norm, while a unique solution might still be retrieved \cite{Chartrand2007,Chartrand2008}.\\
Dictionaries can be classified as parametric or non-parametric.
Non-parametric dictionaries are typically learned from training data and often used if no analytical model is 
\mbox{available \cite{Rubinstein2010}.} 
While they can yield sparser representations of certain data realizations \cite{Mairal2008a}, non-parametric dictionaries 
usually lack an interpretable structure and are inefficient in terms of \mbox{storage \cite{Rubinstein2010}.}
Parametric dictionaries, in turn, rely on an analytical model for the observed signal. %
Their analytic form offers an efficient implementation and a means to obtain %
optimality proofs and error bounds \cite{Rubinstein2010}.
They are also favorable in terms of scalability and storage-\mbox{efficiency \cite{Ataee2010,Rubinstein2010}}.
\emph{Translation-invariant} dictionaries represent an important sub-class of parametric 
dictionaries, that can be used to estimate the translation coefficients of localized \mbox{signals \cite{Jost2006,Blumensath2006,Fyhn2015}.} %
Nonetheless, 
due to the complexity of natural signals, some model parameters might be unknown or contain uncertainty.
Parametric \emph{Dictionary} \emph{Learning} (DL) addresses this problem with the aim of estimating these parameters from the measured data. 
Herein,
statistical DL methods, such as maximum \mbox{likelihood (ML)} or maximum \emph{a posteriori} (MAP) estimation, are commonly employed \cite{Rubinstein2010}.
In order to solve the resulting optimization problem, \emph{alternating estimation} (AE) is a frequently \mbox{pursued sub}-optimal paradigm, that 
iteratively optimizes a local objective \mbox{function \cite{Mohamed2012,Beck2013,Agarwal2014}.} %
In a Bayesian setting, the Expectation Maximization (EM) algorithm is a popular variant of AE-based estimation \cite{Rubinstein2010}.\\
A model for the sparse coefficients can be of deterministic or probabilistic nature. 
While the deterministic case is often assumed in sparse estimation \cite{Donoho2003,Candes2011}, 
a probabilistic model offers high flexibility to take model deviations and measurement errors into account. 
Moreover, a hierarchical structure can be used to incorporate additional uncertainty in prior assumptions. 
Sparsity can either be promoted by continuous distributions, 
resulting in \emph{weakly sparse} models, or by discrete mixtures, leading to \emph{strongly sparse} \mbox{models \cite{Mohamed2012}.} 
A prominent example of discrete mixtures are \emph{Spike \& Slab} models \cite{Ishwaran2005a}. %
They are based on binary activations and yield strongly sparse representations.  
Continuous sparse priors, such as a Gaussian or double-exponential (Laplace) prior, feature high excess kurtosis with heavy tails 
and a narrow peak around \mbox{zero \cite{Polson2010,Mohamed2012}.} %
Besides sparsity, additional knowledge of the signal, e.g. correlation, can be incorporated \cite{Eldar2010,Zhang2011a}.\\  
For many practical models, evaluating the posterior distribution is  not feasible and approximate methods,
such as \emph{Markov Chain Monte Carlo (MCMC)} or variational Bayes methods, have to be used to accomplish 
\mbox{inference \cite{Bishop2006,OHagan1994,Seeger2008}.} 
Variational methods %
use rather simple analytic functions to approximate the posterior distribution by factorization, which is 
favorable in terms of scalability and computational costs but leads to a  
deterministic approximation \cite{Seeger2008,Bishop2006}. 
MCMC methods %
attempt to sample the posterior distribution, where subsequent samples form a \emph{Markov chain} \cite{Bishop2006}. 
The \mbox{\emph{Hamilton Monte Carlo} (HMC)} method is a powerful technique, that is especially suitable %
for sampling high-dimensional spaces in the presence of correlation \cite{Neal2011}.
However, MCMC performance is generally limited by the available computation time, thereby relying on a stochastic approximation. 
Another application of MCMC is found in non-convex optimization, where
Stochastic \mbox{Gradient (SG)} MCMC has gained popularity for large-scale Bayesian learning \cite{Chen2014,Chen2015,Chen2016}.\\ 
In the present work, we consider the problem of \emph{Compressed Fiber Sensing} (CFS) with highly coherent translation-invariant dictionaries and 
imperfectly known parameters. %
For the sparse coefficients, a weakly sparse hierarchical model is considered. %
We also establish a relation between this model %
and non-convex optimization with $\ell_p$-norm constraints for $0\!\!<\!\!p\!\!<\!\!1$. 
In order to alleviate the problem of dictionary coherence, we leverage additional structure of the dictionary and achieve
augmented sparsity %
by establishing a \emph{Markov random} \mbox{\emph{field} (MRF)} relation among the sparse coefficients. 
For dictionary learning, we pursue two different strategies: In the first \mbox{strategy (\textbf{\emph{S1}}),} we consider a deterministic dictionary parameter, that is 
estimated using a Monte Carlo EM algorithm. %
In the second \mbox{strategy (\textbf{\emph{S2}}),} a probabilistic hierarchical model for the dictionary parameter is considered, 
leading to a full Bayesian formulation and joint estimation of the sparse coefficients and the dictionary parameter. %
In both strategies, approximate inference 
is accomplished using a hybrid MCMC method based on Gibbs sampling and HMC. 
Finally, we use simulations and real data to compare the proposed methods to previous work \mbox{in \cite{Weiss2016},} %
where a deterministic model is considered for the sparse coefficients and the dictionary parameter. %
For the deterministic case, we derive the Cram{\'e}r-Rao bound (CRB) to assess the performance gain achieved by a 
probabilistic model.

\subsection{Contributions} %
\begin{itemize}
\item[(I)] We propose a probabilistic model for the sparse coefficients, where a Weibull prior is used to promote (weak) sparsity. %
	  Additional collective shrinkage is achieved by establishing an MRF-relation among the sparse coefficients based on a bivariate kernel 
	  function in the joint prior density.
	  This helps to moderate the impact of severe dictionary coherence and can be used in general sparse synthesis problems with similar dictionary structure. %
	  We also establish a relation to non-convex optimization with constraints on the $\ell_p$-norm for $0<p<1$. 
\item[(II)] For dictionary learning, we investigate two conceptually different strategies, assuming either a deterministic (\textbf{\emph{S1}}) or 
	  a stochastic (\textbf{\emph{S2}}) dictionary parameter. 
	  In both strategies, the noise level can be jointly estimated along with the sparse coefficients. 
	  We further highlight advantages, disadvantages and limitations to offer support in choosing an adequate method for 
	  practical systems. %
\item[(III)] To accomplish inference in these models, we use a hybrid MCMC method, combining HMC and Gibbs sampling,
	  We show its applicability and efficacy in the considered sampling problem for CFS. %
\item[(IV)] We use simulations to evaluate the performance of the proposed sparse estimation and DL methods for various scenarios of 
	  different CS sample sizes, SNRs and CS matrices. %
	  These results are compared to an existing method in \cite{Weiss2016}, where the sparse coefficients and the dictionary parameter 
	  are assumed to be deterministic. In addition, we provide a real-data example to verify the practical applicability of 
	  \textbf{\emph{S1}} and \textbf{\emph{S2}}. %
\item[(V)] We derive the Cram{\'e}r-Rao bound for jointly estimating the sparse coefficients and the dictionary parameter in the deterministic case. It is a valid bound for the 
		    competing method in \cite{Weiss2016}, and serves to assess the achieved performance gain of our probabilistic approach. 
\end{itemize}

\section{Related Work}
\noindent
There exists little work addressing %
the combination of CS and DL for the application of WDM-based distributed fiber-optic \mbox{sensing \cite{Weiss2013,Weiss2015,Weiss2016}.}
In \cite{Weiss2013}, a model for the received sensor signal is presented, from which a redundant shift-invariant parametric dictionary is created.
The works \mbox{in \cite{Weiss2015,Weiss2016}} focus on the aspect of CS and sparse estimation in the case of uncertain dictionary parameters.
The authors use AE-based estimation to determine the dictionary parameters, where a pre-processing routine 
accounts for severe dictionary coherence. 
Unlike our approach, these works use a deterministic model for the sparse coefficients and dictionary parameters.\\ %
Weakly sparse models have been widely used in the literature.
A comprehensive analysis of different %
hierachical sparse prior models is provided in \cite{Mohammad-Djafari2012}. %
The general problem of choosing the prior in weakly sparse models for sparse regression is addressed in \cite{Polson2010}, where 
the authors describe various properties of different shrinkage priors and illuminate the selection problem 
from two perspectives: prior distributions and penalty functions. 
The work in \cite{Mohamed2012} also investigates Bayesian methods with different sparse models in comparison to classical $\ell_1$-minimization.
Seeger \cite{Seeger2008} found that the Laplace prior is able to shrink most components close to zero, while allowing for selected components
to become sufficiently large. 
This effect, termed \emph{selective shrinkage} \mbox{in \cite{Ishwaran2005},}  %
is most noticeable for heavy-tailed priors, e.g. the Student's $t$-prior \cite{Seeger2008} or the \emph{horseshoe} prior in \cite{Carvalho2010,Polson2010}.
Based on these findings, %
we select a sparsity prior that resembles a 
positive version of the horseshoe prior. %
Other works, that focus on the perspective of penalized regression, report higher sparsity levels 
by penalizing the $\ell_p$-norm with $0<p<1$ instead of the $\ell_1$-norm \cite{Gupta2013}. 
The authors in \cite{Chartrand2007} show that the RIP requirements for the dictionary can be relaxed in this case. %
It is also pointed out in \cite{Chartrand2007,Chartrand2008} %
that non-convex CS with $\ell_p$-norm penalization requires less measurements than standard CS, which is based on the $\ell_1$-norm. %
We rely on these results 
and show a relation between the considered sparsity prior and non-convex optimization with $\ell_p$-norm constraints.\\
There exist several approaches to exploit additional structure of the signal. One example is \emph{block sparsity} \cite{Eldar2010}.
A block sparse Bayesian learning framework is proposed in \cite{Zhang2011a}, pointing out how correlation can be exploited in regularization algorithms. 
Wakin \emph{et} \mbox{\emph{al.} \cite{Wakin2005}} introduce the concept of \emph{joint sparsity} for signal recovery in distributed CS theory.
In \cite{Malioutov2005}, temporal correlation across subsequent CS measurements is considered,   %
while the authors in \cite{Chen2013a} use correlation to achieve smoothness. %
Another related concept is proposed in \cite{Altmann2015}, where a truncated multivariate Ising MRF model is used to describe the correlation between adjacent pixels for image processing. 
Different from these works, we use the ideas of MRFs \cite{Murphy2012} and exploit correlation to achieve collective shrinkage among the sparse coefficients.\\ %
A comparative analysis in \cite{Mohamed2012} suggests that MCMC methods are powerful for inference in sparse models.
In \cite{Neal2011}, the benefits of HMC and Gibbs sampling in hierarchical models are outlined. %
It is also shown, that HMC can be more effective than a Gibbs sampler for sampling high-dimensional spaces in the presence of correlation.
According to these results, we consider a hybrid MCMC method that combines HMC and Gibbs sampling for inference in our hierarchical model, where the sparse 
coefficients are high-dimensional and correlated. 
For parametric DL, the Monte Carlo EM algorithm in \textbf{\emph{S1}} represents one variant of the frequently applied AE-based estimation
\mbox{technique \cite{Leigsnering2016,Raja2016}.} %
Comparable to \textbf{\emph{S2}} is the Bayesian framework for sparse estimation and DL in \cite{Hansen2014}.
However, the authors use a Gaussian prior without correlation. 

\subsection{Outline}
\noindent
In \mbox{Section \ref{sec:problem_and_model}}, the signal model for CFS is introduced, and in 
\mbox{Section \ref{sec:CRB}}, the CRB for joint estimation of the deterministic sparse coefficients and dictionary parameters is derived. 
\mbox{Section \ref{sec:weak_sparsity_model}} details the sparsity and local similarity model, while Section \ref{sec:approx_inference} 
describes the hybrid MCMC method for approximate inference in this model. 
The parametric DL strategies \textbf{\emph{S1}} and \textbf{\emph{S2}} are described in Section \ref{sec:PDL}. 
Section \ref{sec:performance} shows the working principle along with a performance analysis of the proposed and an existing 
method based on simulations and experimental data. A discussion of the results and findings is given in Section \ref{sec:discussion}. 
Section \ref{sec:conclusion} concludes this work.

\section{Signal Model} 
\label{sec:problem_and_model}
\noindent
In order to determine the quantity and nature of impairments at the FBGs in a WDM-based fiber sensor, 
the time delays of the reflections from the individual FBGs need to be estimated. 
We adopt the model in \cite{Weiss2013,Weiss2015,Weiss2016}, where
CS-based acquisition is employed to reduce the number of samples to be stored and processed. 
The CS measurements are described by 
\begin{equation}
\label{eq:basic_model}
	\mathbf{y} = \boldsymbol{\Phi}\mathbf{A}(\theta)\mathbf{x} + \mathbf{n}\,,
\end{equation}
where $\boldsymbol{\Phi} \in \mathbb{R}^{M\times L}$ is the CS sampling matrix and
$\mathbf{n}\in \mathbb{R}^{M}$ is a Gaussian noise component with independent and identically distributed (i.i.d.) entries, 
$n_m \sim \mathcal{N}(0,\sigma_n^2)$, $m=1,\dots,M$. 
The vector $\mathbf{x} \in \mathbb{R}^N$ is sparse with $K$ significant components, and   %
$\theta\!\in\!\mathbb{R}$ is a scalar dictionary parameter. %
The matrix $\mathbf{A}(\theta)$ represents a redundant shift-invariant dictionary %
and its columns, %
called \emph{atoms}, represent FBG reflections on a dense grid of delays. 
The indices of the $K$ significant components in $\x$ indicate the desired reflection delays. They are collected in the set $\mathcal{S}\! =\! \{i_1,\dots,i_K\}$. 
We can write the full data likelihood function for this model by 
\begin{equation}
\label{eq:Gauss_likelihood}
\hspace{0.05cm}	\p{\y}{\x,\theta} = (\sqrt{2\pi}\sigma_n)^{-M}\exp{\!-\frac{1}{2\sigma_n^2}\|\y-\boldsymbol{\Phi}\mathbf{A}(\theta)\x\|_2^2}\!.\hspace{-0.3cm}
\end{equation}
The $i$-th dictionary atom, $i\!=\!1,\dots,\!N\!$, is defined \mbox{by \cite{Weiss2016}}
\begin{equation}
\label{eq:dict_atoms_elements}
	[\mathbf{a}_i]_l(\theta) = r(lT_d - i\delta t, \theta),\ \ l=1,\dots,L 	 \,,
\end{equation}
where the generating function, $r(lT_d - i\delta t, \theta)$, describes the reflection from a single FBG, 
incrementally shifted by $\delta t$ and sampled with a design sampling period, $T_d$. 
In order to specify the dictionary parameter in CFS according to \cite{Weiss2016}, we write  
\begin{equation}
\label{eq:sensor_signa_IFT_model}
	r(t,\theta) = \int_{-\infty}^{\infty} \text{e}^{j2\pi f t} H_{\text{\tiny LP}}(f,\theta)\,i_{\text{ph}}(f)\, \text{d}f     \,. 
\end{equation}
Herein, $i_{\text{ph}}(f)$ is the received photocurrent in the frequency domain, and 
$H_{\text{\tiny LP}}(f,\theta)$ is the transfer function of a 
lowpass filter, that models a limited \emph{effective bandwidth} of the receiver circuitry. %
This bandwidth is described in terms of a positive dictionary parameter, $\theta\in\mathbb{R}_+$. %
As an auxiliary parameter, it accounts for different indistinguishable sources of uncertainty, that all contribute to the broadening in the temporal response of the FBG
reflections. 
A detailed model for $i_{\text{ph}}(f)$ is \mbox{provided in \cite{Weiss2016}.}

\section{The CRB for joint estimation of ($\x,\theta$) in CFS}  
\label{sec:CRB}
\noindent %
We derive the CRB for jointly estimating the deterministic parameters ($\x,\theta$). This is a valid bound for the model considered 
in \cite{Weiss2016} and can be used to assess the 
relative performance gain achieved by the proposed probabilistic sparse model and DL strategies.
Although the Bayesian CRB in \cite{Bobrovsky1987} can be empirically determined, we found that this bound is very lose, due to the high information content in the considered sparsity prior. %
Therefore, and in regard of the comparative analysis with the deterministic case in \mbox{\cite{Weiss2016}}, the non-Bayesian 
CRB is more useful in this case.\\  %
The constrained CRB for estimating $\x$ with sparsity constraints has been derived in \cite{Ben-Haim2010}.
However, this derivation does not assume uncertainty in the dictionary. %
It is based on locally balanced sets and involves the projection of the \emph{Fisher Information matrix} (FIM), $\boldsymbol{\mathcal{I}}(\x)$,
onto a low-dimensional subspace spanned by the so-called \emph{feasible directions}. Any estimator, $\hat{\x}$, for which the constrained CRB is 
a valid lower bound, must be unbiased 
with respect to these directions. The projection matrix can be created from the unit vectors corresponding to the 
non-zero coefficients in $\x$, that is $\mathbf{U}=[\mathbf{e}_{i_1},\dots,\mathbf{e}_{i_K}]$ with $i_k\in \mathcal{S}$, 
\mbox{$k=1,\dots,K$.} 
For a Gaussian likelihood as in (\ref{eq:Gauss_likelihood}), the FIM can be derived 
from the expected value of the Hessian matrix of the log-likelihood function, i.e. \cite{Kay1993,Ben-Haim2010}
\begin{equation}
	\boldsymbol{\mathcal{I}}(\x) = - \EXP{\y}{}\grad{\x}^2 \log \p{\y}{\x,\theta} = \frac{1}{\sigma_n^2}\mathbf{B}^{\!\top}\mathbf{B},
\end{equation}
with $\mathbf{B} = \boldsymbol{\PHI}\mathbf{A}$.
Further, we define the reduced FIM by \mbox{$\boldsymbol{\mathcal{I}}_K := \mathbf{U}^{\!\top}\boldsymbol{\mathcal{I}}(\x)\mathbf{U}$}.
Then, given that $\x$ is exactly $K$-sparse, the constrained CRB for a \emph{known} dictionary becomes \cite{Ben-Haim2010}
\begin{equation}
	\text{Cov}(\hat{\x}) \succeq \mathbf{U}\,\boldsymbol{\mathcal{I}}_K^{-1}\mathbf{U}^{\!\top},\quad  \|\x\|_0 = K.
\end{equation}
Based on these results, we derive the CRB for the joint parameters $\boldsymbol{\gamma} = (\x,\theta)$.
First, we derive the Fisher information for $\theta$, given that $\x$ is known. It is given by
\begin{eqnarray}
\nonumber	\mathcal{I}(\theta) &=& - \EXP{\y}{} \partialfracsq{}{\theta}\log \p{\y}{\x,\theta}\\[0.0cm]
\nonumber 	  					&=&   \EXP{\y}{} \partialfracsq{}{\theta}\frac{1}{2\sigma_n^2} (\y-\PHI\mathbf{A}(\theta)\x)^{\!\top} ( \y-\PHI\mathbf{A}(\theta)\x)\\[0.0cm]
\label{eq:Fisher_theta} 	    &=& \frac{1}{\sigma_n^2}\,\x^{\!\top}\mathbf{A}'(\theta)^{\!\top}\PHI^{\!\top}\PHI\mathbf{A}'(\theta)\x .
\end{eqnarray}
Herein, $\mathbf{A}'(\theta)$ denotes the (element-wise) derivative of $\mathbf{A}(\theta)$ with respect to $\theta$. 
Next, we have to take into account that $\x$ and $\theta$ share some mutual information. Therefore, we define the combined FIM: %
\begin{equation}
	\boldsymbol{\mathcal{I}}(\boldsymbol{\gamma}) = \left(
	\begin{matrix}
		\boldsymbol{\mathcal{I}}(\x) & -\EXP{\y}{}\mathbf{u}\\[0.1cm]
		-\EXP{\y}{}\mathbf{u}^{\!\top}    			 & \mathcal{I}(\theta)	
	\end{matrix} \right),
\end{equation}
where $\mathbf{u}=[u_1,\dots,u_N]^T$ and $u_i = \partialfrac{}{x_i}\partialfrac{}{\theta}\log \p{\y}{\x,\theta}$, \mbox{$i=1,\dots,N$.}
Since the partial derivatives can be interchanged, the off-diagonal elements are identical. 
In order to complete the definition of $\boldsymbol{\mathcal{I}}(\boldsymbol{\gamma})$, we determine
\begin{eqnarray}
\nonumber -\EXP{\y}{\!u_i} &=& -\EXP{\y}{}\frac{\partial^2}{\partial x_i\partial\theta}  \log\p{\y}{\x,\theta}\\
				  &=& \frac{1}{\sigma_n^2}\x^{\!\top}\mathbf{A}'(\theta)^{\!\top}\PHI^{\!\top}\PHI\,\mathbf{a}_i(\theta).
\end{eqnarray}
The reduced FIM is obtained by appending the set of feasible directions, such that the coordinate  $\theta$ is included, i.e. 
\mbox{$\mathbf{\tilde{U}} = [\mathbf{U},\mathbf{e}_{N+1}]$.} Hence, $\boldsymbol{\mathcal{I}}_{K+1}=\mathbf{\tilde{U}}^{\!\top}\boldsymbol{\mathcal{I}}(\boldsymbol{\gamma})\mathbf{\tilde{U}}$.
To obtain the inverse, we apply twice the matrix inversion lemma \cite{Higham2002}    %
\begin{equation}
	\boldsymbol{\mathcal{I}}_{K+1}^{-1} = 
	\left(\begin{matrix}
		\left(\boldsymbol{\mathcal{I}}_K-\frac{\mathbf{v}\mathbf{v}^{\!\top}}{\mathcal{I}(\theta)}\right)^{-1} & 
			-\frac{1}{\breve{b}}\boldsymbol{\mathcal{I}}_K^{-1}\mathbf{v} \\[0.2cm]
	    -\frac{1}{\breve{b}}\mathbf{v}^{\!\top}\boldsymbol{\mathcal{I}}_K^{-1} & 
	    	\frac{1}{\breve{b}}
		 \end{matrix}\right),
\end{equation}
where $\breve{b} = \mathcal{I}(\theta)-\mathbf{v}^{\!\top}\boldsymbol{\mathcal{I}}_K^{-1}\mathbf{v}$, and  
\begin{equation}
	\left(\boldsymbol{\mathcal{I}}_K-\frac{\mathbf{v}\mathbf{v}^{\!\top}}{\mathcal{I}(\theta)}\right)^{-1} = \boldsymbol{\mathcal{I}}_K^{-1}+\frac{1}{\breve{b}}\boldsymbol{\mathcal{I}}_K^{-1}\mathbf{v}\mathbf{v}^{\!\top}\boldsymbol{\mathcal{I}}_K^{-1}\,.
\end{equation}
The constrained CRB for the joint parameters in $\boldsymbol{\gamma}$ becomes
\begin{equation}
\text{Cov}(\boldsymbol{\gamma})\ \succeq\ \mathbf{\tilde{U}}\boldsymbol{\mathcal{I}}_{K+1}^{-1}\mathbf{\tilde{U}}^{\!\top},\quad \|\x\|_0 = K\,.
\end{equation}
Finally, a lower bound for the \emph{mean squared error (MSE)} in the joint setting is obtained by updating the individual estimation errors to
account for the information shared between $\x$ and $\theta$:  %
\begin{eqnarray}
\hspace{-0.4cm}	\MSE{\hat{\x}} &\!\!\! \geq&\!\!\! (\trace{\boldsymbol{\mathcal{I}}_{K}^{-1}}) + \frac{1}{\breve{b}} \mathbf{v}^{\!\top}\boldsymbol{\mathcal{I}}_{K}^{-1}\boldsymbol{\mathcal{I}}_{K}^{-1}\mathbf{v}\\[0.1cm] %
\hspace{-0.4cm}	\MSE{\hat{\theta}}&\!\!\! \geq&\!\!\! \frac{1}{\breve{b}}\, =\ \mathcal{I}(\theta)^{-1} + \frac{\mathbf{v}^{\!\top}\boldsymbol{\mathcal{I}}_K\mathbf{v}}{\mathcal{I}(\theta)\,(\,\mathcal{I}(\theta)-\mathbf{v}^{\!\top}\boldsymbol{\mathcal{I}}_K\mathbf{v})}.
\end{eqnarray}

\section{Probabilistic sparse model} %
\label{sec:weak_sparsity_model}
\noindent
Regarding the model in (\ref{eq:basic_model}), the data can be explained in different ways. 
On the one hand, many non-zero components in $\mathbf{x}$ and a large bandwidth, $\theta$, result in many narrow temporal peaks that can yield a good approximation of the observed reflections.
On the other hand, it is known that the sensing fiber contains $K$ FBGs, so we expect exactly $K$ reflections. 
Therefore, a more useful explanation is given by $K$ significant elements in $\mathbf{x}$ with a smaller value of $\theta$, such that  
$\mathcal{S}$ correctly indicates the reflection delays. 
Nevertheless, even for a suitable value of $\theta$, the signal $\mathbf{x}$ is usually not exactly sparse but contains many small elements
close to zero, e.g. due to measurement noise. In a strongly sparse model, these contributions are not taken into account, which
impacts the positions of non-zero elements in $\x$. Hence, it may lead to incorrectly estimated reflection delays. %
This motivates a weakly sparse model, where the $K$ most significant components indicate the reflection delays.  
When $\x$ and $\theta$ are both unknown, the reflections delays can only be estimated when prior information of sparsity is incorporated,
since $\theta$ depends on $\mathbf{x}$ and vice versa. %
Severe dictionary coherence aggravates this problem and 
results in several non-zero components with moderate amplitudes around the true significant elements. 
The coherence level is even stronger when the dimensionality of the acquired data is further reduced by CS. 
Thus, an adequate sparse model for $\x$ must compensate for this effect. %
Classic $\ell_1$-minimization can be interpreted as an MAP estimation problem, where $\mathbf{x}$ has i.i.d. entries 
with Laplace \mbox{priors \cite{Mohamed2012}.} %
However, the required performance guarantees for $\ell_1$-minimization, essentially the RIP \cite{Candes2005,Candes2008a},
are no longer fulfilled in the case of strong dictionary coherence. 
According to \cite{Chartrand2007,Chartrand2008}, the RIP conditions can be relaxed for $\ell_p$-minimization, when $0<p<1$.
Therefore, we use a prior with stronger selective shrinkage effect, that can be related to constraints on 
the $\ell_p$-norm in non-convex optimization.
Yet, specific characteristics of the signal have to be considered. %
The measured reflection signal is proportional to the optical power, and the dictionary atoms essentially model the optical power reflected 
from the individual FBGs. Thus, the prior must also account for the non-negativity of the data.
Due to these restrictions, we choose a Weibull prior that resembles a positive version of the horseshoe prior \mbox{in \cite{Polson2010}} 
and induces the required selective shrinkage effect:
\begin{equation}
\hspace{0.25cm}	x_i \sim p(x_i) = \Weibull{x_i}{\lWeibull,\kWeibull}\,, \  x_i > 0,  \  i=1,\dots,N, \hspace{-0.3cm}
\end{equation} 
where $\!\lambda_w,k_w$ are the scale and shape parameters, respectively. 
Then, the joint prior density of $\mathbf{x}$ is given by
\begin{equation}
\label{eq:joint_x_Weibull}
	\p{\mathbf{x}}{\kWeibull,\lWeibull} = \frac{\kWeibull}{\lWeibull^{\kWeibull}} \prod_{i=1}^{N} x_i^{\kWeibull-1}\,\exp{\!-\lWeibull^{-\kWeibull} \sum_{i=1}^{N}x_i^{\kWeibull}\!}\!.\hspace{-0.1cm}
\end{equation}
Fig. \ref{fig:kernel_and_impact_on_pdf} (top left)
shows qualitatively the shape of the considered prior in the bivariate case.\\
Based on (\ref{eq:joint_x_Weibull}) and (\ref{eq:Gauss_likelihood}), we can relate the problem to 
constrained ML estimation.
First, let us consider an interpretation in terms of MAP estimation as in \cite{Mohammad-Djafari2012}, by 
calculating $\text{arg}\max_{\x}\, \log\,\p{\y}{\x,\theta}\p{\mathbf{x}}{\kWeibull,\lWeibull}$ or, equivalently, 
\begin{eqnarray}
\nonumber\hspace{-1.2cm}	&&\hspace{-0.0cm}	\text{arg}\min_{\hspace{-0.3cm}\x}\ -\log\,\p{\y}{\x,\theta}\p{\mathbf{x}}{\kWeibull,\lWeibull}\ = \\
\hspace{-1.2cm}	&&\hspace{-0.0cm} \text{arg}\min_{\hspace{-0.3cm}\x}\, \|\y\!-\!\boldsymbol{\Phi}\mathbf{A}(\theta)\x\|_2^2 
+ \mu_1\!\sum_{i=1}^N\log(x_i)  + \mu_2\!\sum_{i=1}^{N}x_i^{\kWeibull}\!,
\end{eqnarray}
where $\mu_1=(1-\kWeibull)$ and $\mu_2=\lWeibull^{-\kWeibull}$ with \mbox{$0<\kWeibull <1$} and $\mu_1,\mu_2 >0$.
In order to formulate a related constrained ML problem,
let us define two functions, 
\begin{equation}
\label{eq:constr_fcts}
	g_1 = \sum_{i=1}^{N}x_i^{\kWeibull}-\lambda_1^{\kWeibull}\quad\ \  \text{and}\quad\ \ g_2 = \sum_{i=1}^N\log(x_i)-\lambda_2,
\end{equation}
where $\lambda_1,\lambda_2 \in \mathbb{R}_+$ are related to the coefficients 
$\mu_1,\mu_2$, respectively.
The functions in (\ref{eq:constr_fcts}) can represent inequality constraints of the form $g_1 \leq 0$ and $g_2\leq 0$, 
that account for the impact of the prior by restricting the search  %
space. \mbox{Hence, a} constrained version of the ML problem can be formulated by   
\begin{eqnarray}
\label{eq:optProblem_costFct}
	\hspace{0.1cm}\text{arg}\min_{\hspace{-0.3cm}\x \succ \mathbf{0}} &\ \ \,   \|\y-\boldsymbol{\Phi}\mathbf{A}(\theta)\x\|_2^2&\\[0.0cm]
\label{eq:optProblem_constr_1}
	\text{s.t.}& \ \ \|\mathbf{x}\|_{k_w}   &\leq\ \lambda_1 \\[0.1cm]  %
\label{eq:optProblem_constr_2}
	\text{and}&  \ \ \sum_{i=1}^{N}\log(x_i)& \leq\ \lambda_2 \,.
\end{eqnarray}
In this non-convex problem, $\|\mathbf{x}\|_p = (\sum_{i=1}^{N}|x_i|^p)^{1/p}$ denotes the $\ell_p$-norm with \mbox{$p = \kWeibull < 1$}.
The hyperparameters $\lambda_1, \lambda_2$ control the shrinkage effects. 
Fig. \ref{fig:kernel_and_impact_on_pdf} (top right)
depicts the search space restricted by the constraints (\ref{eq:optProblem_constr_1})-(\ref{eq:optProblem_constr_2}).
\mbox{The borders} are shown for a fixed value of $\lambda_1$ and $\lambda_2$ in the bivariate case.

\subsection{Local covariance model for augmented sparsity}
\begin{figure}[t]
\centering
	\includegraphics[width=0.63\columnwidth]{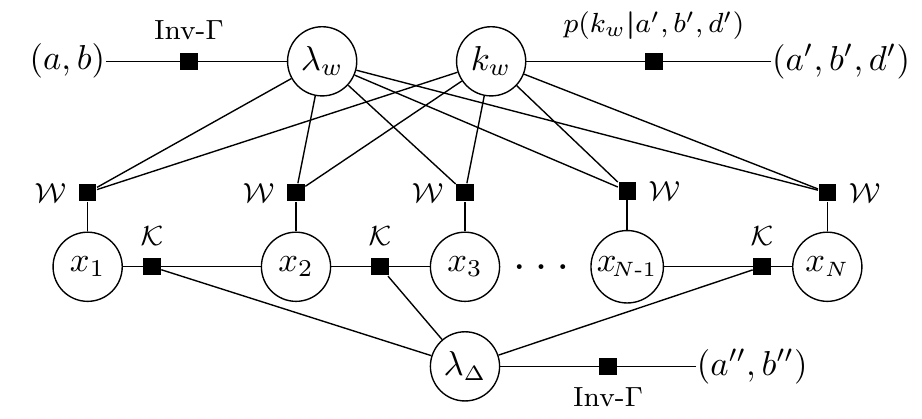}
\caption{Factor graph of the complete sparse model with local similarity.}
\label{fig:factor_graph_localSimilarity} 
\end{figure}

\noindent
In analogy to the concept of block sparsity \cite{Eldar2010}, we can use
the specific sparse structure of the signal with respect to the shift-invariant dictionary for CFS
to exploit sparsity among groups of variables. %
The signal contains only $K$ reflections that arrive at temporally separated delays, indicated by the significant components in $\x$.
Therefore, we can assume that a significant coefficient is always surrounded by larger groups of non-significant coefficients and 
any two significant components are always well separated. %
Also, it is likely that the amplitudes of adjacent non-significant coefficients are similarly close to zero. %
Borrowing from the ideas of MRFs \cite{Murphy2012}, such local similarity can be modeled by a prior on the differential coefficients, $\Delta\x$, 
where $\Delta x_i = x_{i+1} - x_{i},\, i=1,\dots,N-1$. It restricts the variation of adjacent amplitudes and establishes a MRF relation between
neighboring coefficients in $\x$.
Then, non-significant coefficients with larger amplitudes are pulled down to match the majority with almost-zero amplitudes, 
which promotes additional \emph{collective} shrinkage. 
However, if a significant coefficient follows a non-significant one (or vice versa), the model should allow for larger changes. 
Therefore, the differential variation must be locally specified, dependent on the respective amplitudes, in order to avoid undesired 
shrinkage or equalization.  %
\begin{figure}[t] 
\centering
\subfigure[]{\label{4figs-a}\includegraphics[width=0.34\columnwidth]{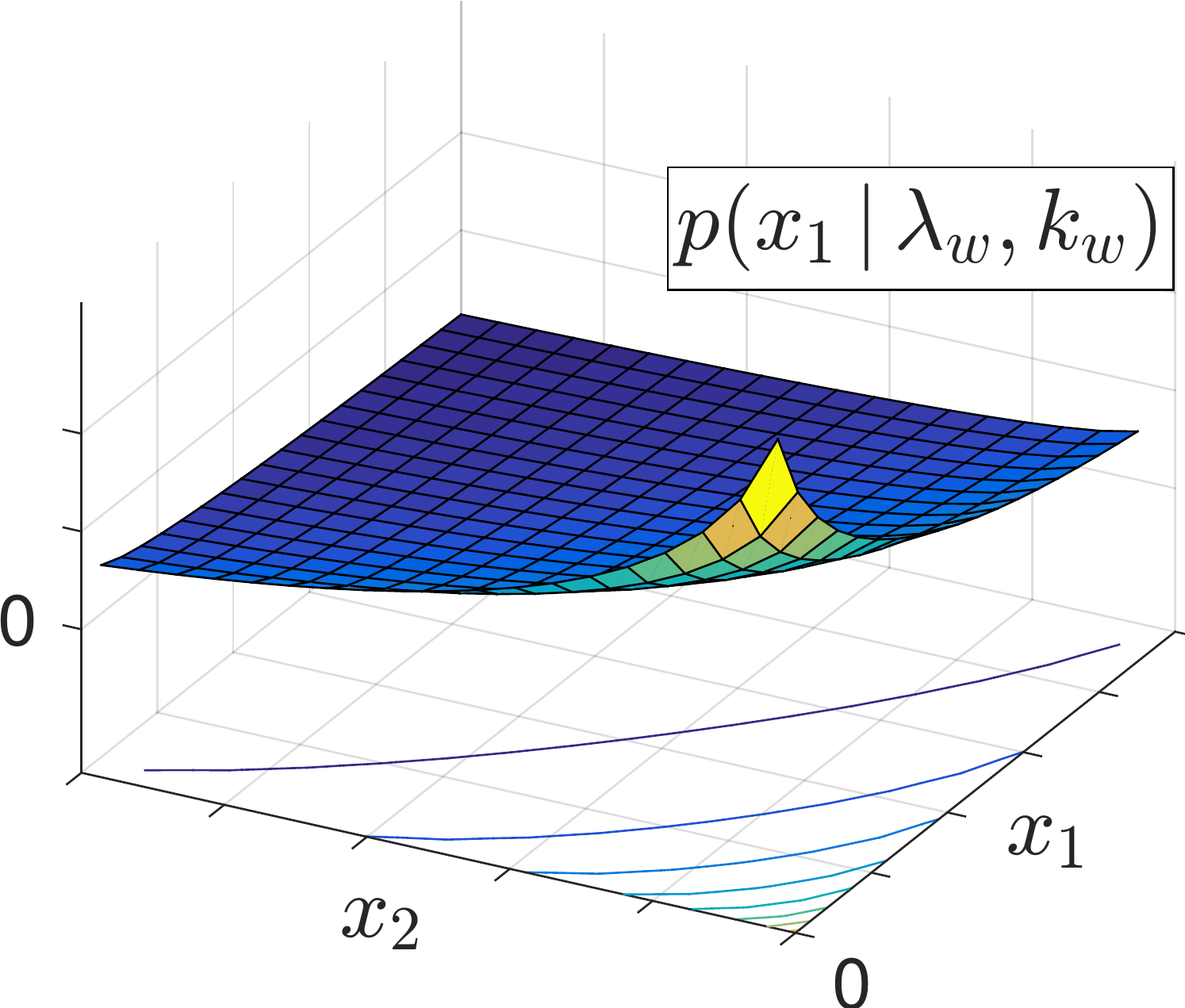}}\hspace{2.0cm}
\subfigure[]{\label{4figs-b}\includegraphics[width=0.30\columnwidth]{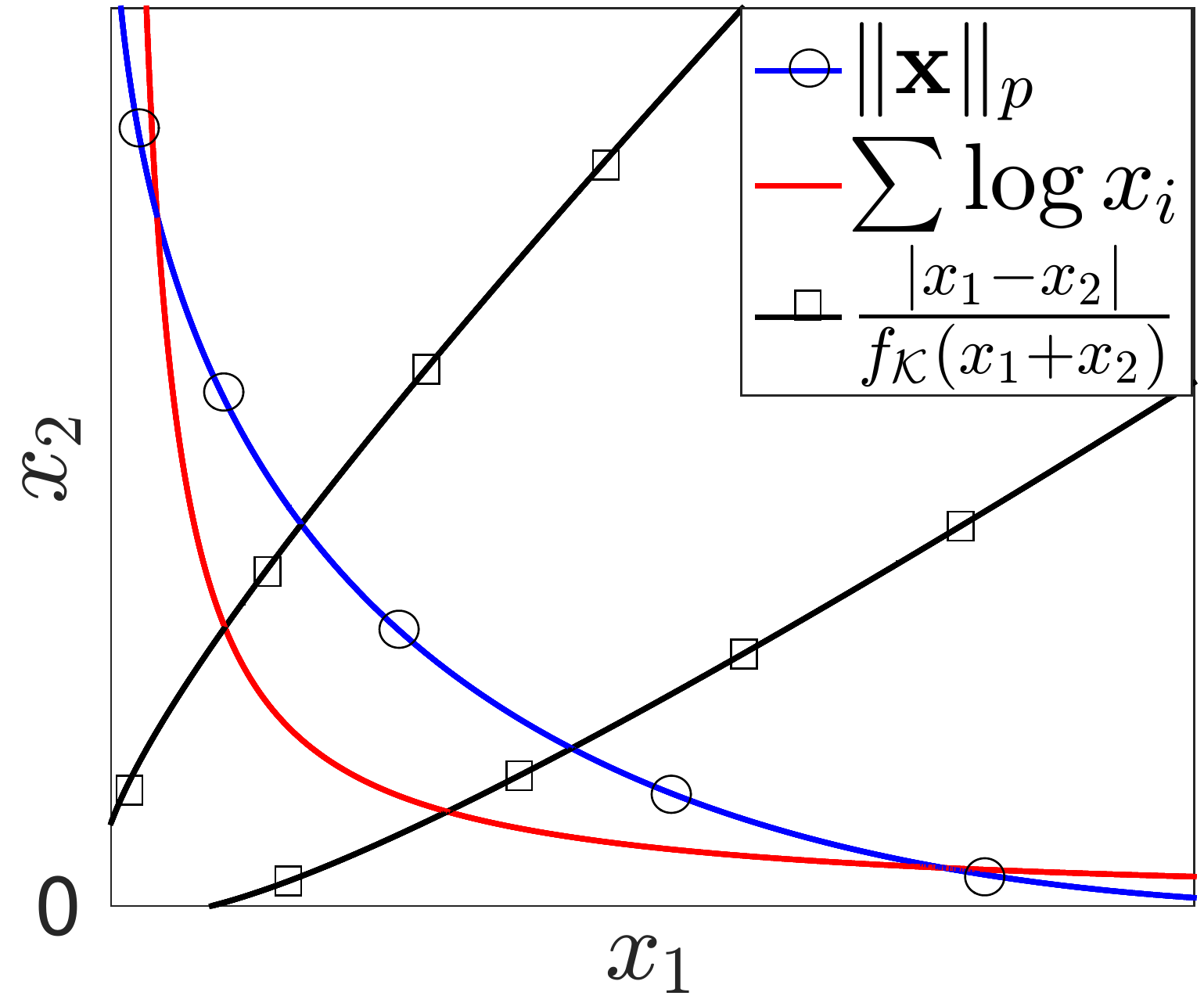}}\\[0.5cm]
\subfigure[]{\label{4figs-c}\includegraphics[width=0.34\columnwidth]{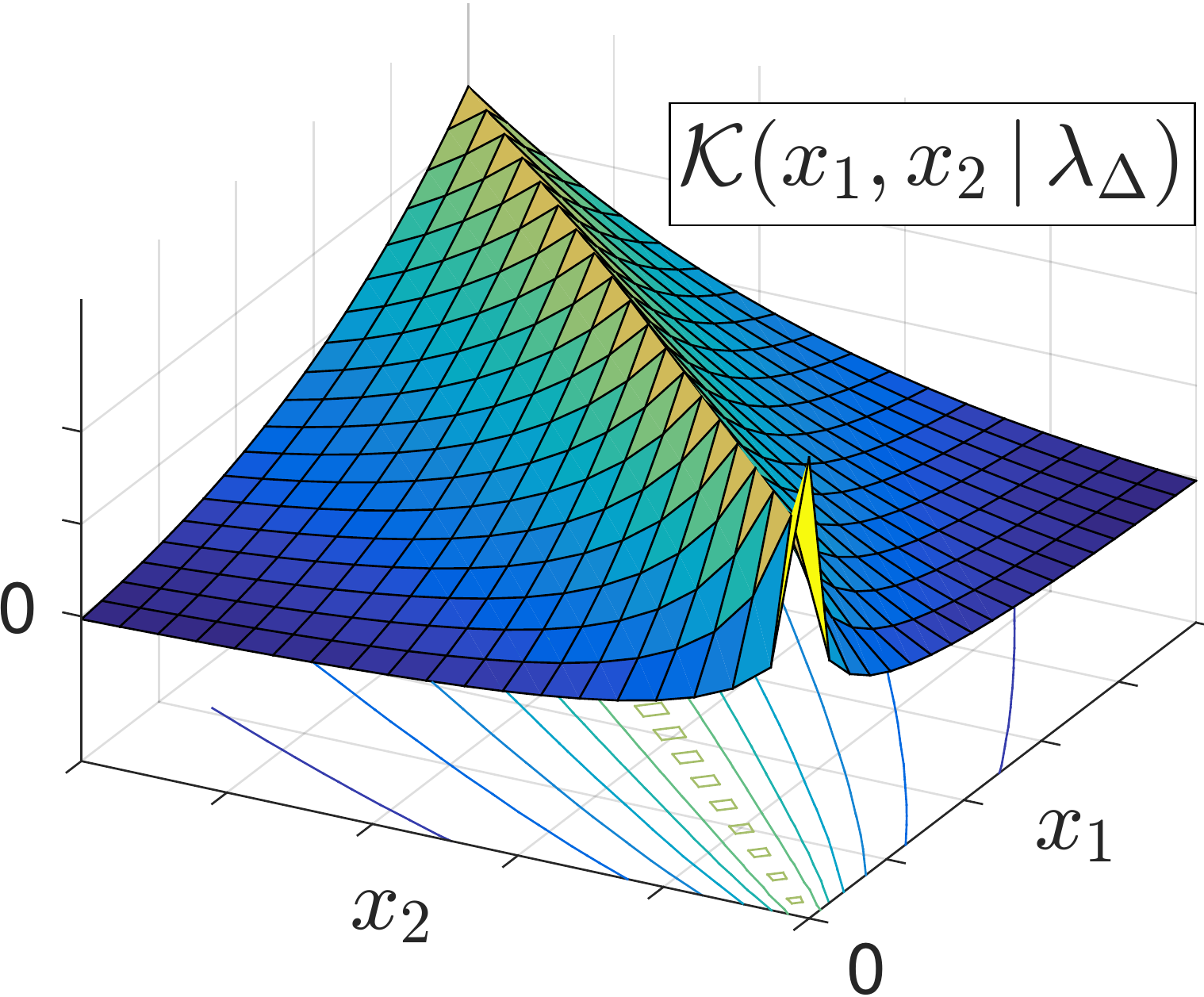}}\hspace{2.0cm}
\subfigure[]{\label{4figs-d}\includegraphics[width=0.34\columnwidth]{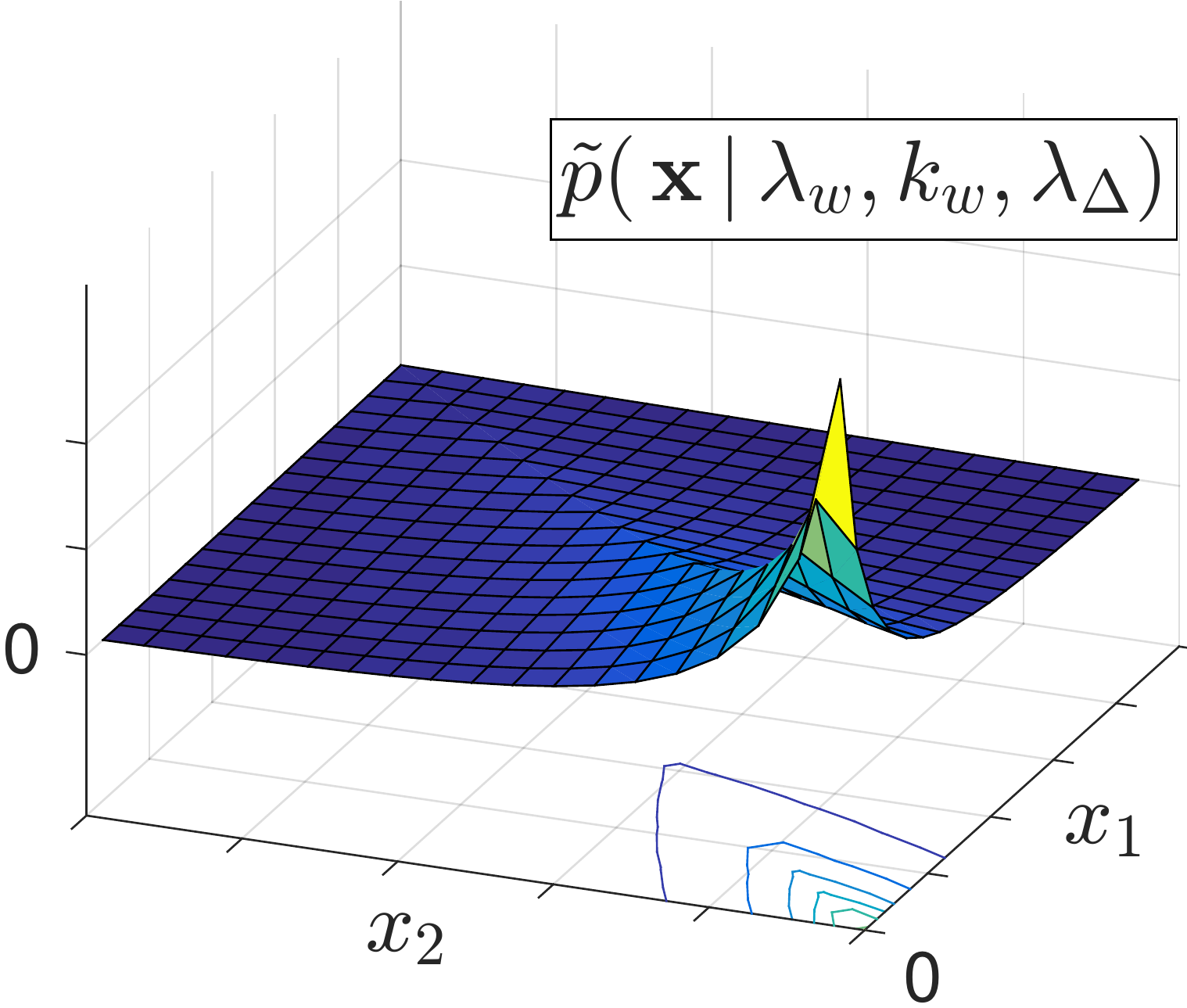}}%
\caption{Bivariate examples. Top left: Weibull prior 
	$p(x_1\,\boldsymbol{|}\,\lWeibull,\kWeibull)$. 
	Top right: search space of the constrained non-convex optimization problem.
	Bottom left: local similarity function 
	$\mathcal{K}(x_1,\!x_2\boldsymbol{|}\lambda_{\text{\tiny$\Delta$}})$. 
	Bottom right: modified joint density $\tilde{p}(\x\boldsymbol{|}\kWeibull,\!\lWeibull,\!\lambda_{\text{\tiny$\Delta$}}\!)$ 
	with $\x\!\!=\!\![x_1,\!x_2]^{\!\top}$\!\!\!\! 
	and $f_{\text{\tiny$\mathcal{K}$}}(x_1, x_{2})\!=\!\sqrt{\!x_1\!\!+\!x_2}$.}
\label{fig:kernel_and_impact_on_pdf}
\end{figure}
To this end, we define a kernel function for all adjacent pairs of sparse coefficients, i.e. $\forall\ i=1,\dots,N-1$, with hyperparameter $\lambda_{\text{\tiny$\Delta$}}$:
\begin{equation}
\label{eq:kernel}
	\mathcal{K}(x_{i},x_{i+1}\,\boldsymbol{|}\,\lambda_{\text{\tiny$\Delta$}})\ =\ \exp{-\lambda_{\text{\tiny$\Delta$}}\frac{|x_{i+1}-x_{i}|}{f_{\text{\tiny$\mathcal{K}$}}(x_i,x_{i+1})}}\,.\vspace{0.05cm}
\end{equation}
The bivariate function $\!f_{\text{\tiny$\mathcal{K}$}}$ %
controls the similarity level between adjacent coefficients. 
Within the scope of this work we consider cases, where this function takes the form \mbox{$f_{\text{\tiny$\mathcal{K}$}}(x_i,x_{i+1}) = (x_i + x_{i+1})^r/N_x$,} $i=1,\dots,N-1$, with positive constants $r\leq 1$, 
$N_x < \infty$. 
They can be incorporated in $\p{\!\x}{k_w,\!\lambda_w\!}$ to yield a modified joint prior density, 
$$\hspace{-2.3cm}\tilde{p}(\mathbf{x}\,\boldsymbol{|}\,\kWeibull,\lWeibull,\lambda_{\text{\tiny$\Delta$}})\ \ =\ \ \frac{1}{Z_{\text{\tiny$\mathcal{K}$}}} \Weibull{x_N}{\kWeibull,\lWeibull}$$ \vspace*{-0.4cm}
\begin{equation}
  \label{eq:modified_joint_x} %
\hspace{2.417cm}	\times \prod_{i=1}^{N-1}\!\mathcal{K}(x_i,x_{i+1}\,\boldsymbol{|}\,\lambda_{\text{\tiny$\Delta$}})\, \Weibull{x_i}{\kWeibull,\lWeibull},\hspace{-0.6cm}
\end{equation}
with normalization constant $Z_{\text{\tiny$\mathcal{K}$}}$. %
For any $\alpha,\beta\in \mathbb{R}_+$, %
it holds that %
$0<\mathcal{K}(\alpha,\beta\,\boldsymbol{|}\,\lambda_{\text{\tiny$\Delta$}}) = \mathcal{K}(\beta,\alpha\,\boldsymbol{|}\,\lambda_{\text{\tiny$\Delta$}}) \leq 1$ and
\begin{eqnarray}
\hspace{-0.5cm}   \left.\tilde{p}(\mathbf{x}\,\boldsymbol{|}\,\kWeibull,\lWeibull,\lambda_{\text{\tiny$\Delta$}})\right|_{Z_{\text{\tiny$\mathcal{K}$}}=1} 
   			\!\!&\leq&\!   \p{\mathbf{x}}{\kWeibull,\lWeibull}  
\end{eqnarray}
is bounded. Hence,  
there exists a positive constant \mbox{$Z_{\text{\tiny$\mathcal{K}$}} < \infty$} that normalizes (\ref{eq:modified_joint_x}) to make $\tilde{p}(\mathbf{x}\,\boldsymbol{|}\,\kWeibull,\lWeibull,\lambda_{\text{\tiny$\Delta$}})$ a proper density. 
Fig. \ref{fig:kernel_and_impact_on_pdf} (bottom) visualizes the function $\mathcal{K}(x_{i},x_{i+1}\,\boldsymbol{|}\,\lambda_{\text{\tiny$\Delta$}})$ and its impact on the original prior in the bivariate case.\\
In the view of constraint ML estimation, the modified prior density in (\ref{eq:modified_joint_x}) can be related to 
the optimization problem %
in (\ref{eq:optProblem_costFct})-(\ref{eq:optProblem_constr_2}) by imposing 
additional constraints
\begin{equation}
	\frac{|x_{i+1}-x_{i}|}{f_{\text{\tiny$\mathcal{K}$}}(x_i,x_{i+1})} \leq \mu_i,\quad i=1,\dots,N-1\,.
\end{equation}
Fig. \ref{fig:kernel_and_impact_on_pdf} (top right) depicts a bivariate example. 
In order to show the MRF relation between the coefficients, 
we calculate  
the conditional densities $\forall\ x_i,\, i=1,\dots,N$.
To this end, we conveniently define \mbox{$\p{\!x_i}{\mathbf{x}_{\setminus i}\!}\! =\! \p{\!x_i}{x_1,\dots,x_{i\text{-}1},x_{i\text{+}1},\dots,x_N\!}$} and get
$$\hspace{-0.5cm}\tilde{p}(x_i\,\boldsymbol{|}\,\mathbf{x}_{\setminus i},\kWeibull,\lWeibull,\lambda_{\text{\tiny$\Delta$}})\ =\ \,
 		\tilde{p}(x_i\,\boldsymbol{|}\,x_{i-1},x_{i+1},\kWeibull,\lWeibull,\lambda_{\text{\tiny$\Delta$}}) 				$$\\[-0.9cm]
\begin{equation}
\label{eq:px_i_cond}
\hspace{0.4cm}\propto\ \Weibull{x_i}{\kWeibull,\lWeibull}\,\mathcal{K}(x_{i-1},x_{i}\,\boldsymbol{|}\,\lambda_{\text{\tiny$\Delta$}})\,\mathcal{K}
				(x_{i},x_{i+1}\,\boldsymbol{|}\,\lambda_{\text{\tiny$\Delta$}}),\hspace{-0.1cm}
\end{equation}\vspace{-0.3cm}
\begin{equation}
\label{eq:px_1_cond}
\hspace{-0.80cm}   	\tilde{p}(x_1\boldsymbol{|}\mathbf{x}_{\setminus 1},\!\kWeibull,\!\lWeibull,\!\lambda_{\text{\tiny$\Delta$}}\!)\ \ \ \,	\propto\ \Weibull{\!x_1\!}{\kWeibull,\!\lWeibull\!}\,\mathcal{K}(x_{1},\!x_{2}\boldsymbol{|}\lambda_{\text{\tiny$\Delta$}}\!),
\end{equation}\vspace{-0.3cm}
\begin{equation}
\label{eq:px_N_cond}
\hspace{-0.75cm}   	\tilde{p}(x_N\boldsymbol{|}\mathbf{x}_{\setminus N}\!,\kWeibull,\!\lWeibull,\!\lambda_{\text{\tiny$\Delta$}}\!)\ \,	\propto 
\Weibull{\!x_N\!}{\kWeibull,\!\lWeibull\!}\,\mathcal{K}(x_{N\text{-}1},\!x_{N}\boldsymbol{|}\lambda_{\text{\tiny$\Delta$}}\!)\hspace{-0.01cm}, 
\end{equation}
where dependencies appear only between directly adjacent coefficients.\\
In order to account for deviations from prior assumptions, we consider randomization of the hyperparameters and assign conjugate %
inverse Gamma priors to the scale parameters $\lambda_w$ and $\lambda_{\text{\tiny$\Delta$}}$.
Finally, given $\lambda_w$ and a normalization constant $Z_{\kWeibull}$, the shape parameter, $\kWeibull > 0$, is assigned the conjugate prior distribution according to \cite{Fink1997}:
\begin{equation}
\label{eq:kw_prior}
\hspace{0.05cm}	\p{\kWeibull}{a',b',(d\,')^{k_w}\!\!,\lWeibull} = \frac{\kWeibull^{a'}}{Z_{\kWeibull}} \exp{\!\!-b'\kWeibull\! - \frac{(d\,')^{\kWeibull}}{\lWeibull}\!}, \hspace{-0.15cm}
\end{equation}
Fig. \ref{fig:factor_graph_localSimilarity} shows a factor graph for the complete sparsity model with %
randomized hyperparameters. %
\section{Approximate Inference: Hybrid MCMC} %
\label{sec:approx_inference}
\noindent
In order to accomplish inference in the sparse model, we apply a hybrid MCMC technique, i.e. HMC within Gibbs sampling.
The reasons for using HMC are twofold: Firstly, it only requires an analytic expression for the posterior density to be sampled. 
Secondly, it is efficient in sampling high-dimensional spaces in the presence of correlation.  %
However, as pointed out in \cite{Neal2011}, it can be more efficient to sample the hyperparameters separately, as their posterior 
distributions are often highly peaked and require a small step size in the HMC algorithm, which limits the general performance. 
Therefore, we employ an outer Gibbs sampler for approximate inference of the latent variables. In each iteration, %
$\tilde{p}(\x\,\boldsymbol{|}\,\lWeibull,\kWeibull,\lambda_{\text{\tiny$\Delta$}})$ is sampled using HMC, while all other variables are fixed. 
Since we are also interested in estimating the noise variance, $\sigma_n^2$, it is 
assigned an inverse Gamma ($\text{Inv-}\Gamma$) prior and sampled along with the other variables. %
The resulting model is summarized below:
\begin{eqnarray}
\nonumber \mathbf{x}\,\boldsymbol{|}\,\kWeibull,\lWeibull,\lambda_{\text{\tiny$\Delta$}}&\sim& \tilde{p}(\x\,\boldsymbol{|}\,\kWeibull,\lWeibull,\lambda_{\text{\tiny$\Delta$}})\,\ \ \quad\qquad \; \text{in}\ (\ref{eq:modified_joint_x}),\\
\nonumber \lWeibull &\sim&	\text{Inv-}\Gamma(\lWeibull\,\boldsymbol{|}\,a,b),\\
\nonumber \kWeibull\,\boldsymbol{|}\,\lWeibull &\sim&	\p{\kWeibull}{a',b',(d\,')^{k_w},\lWeibull}\; \; \; \;  \text{in}\ (\ref{eq:kw_prior}),\\
\nonumber \lambda_{\text{\tiny$\Delta$}} &\sim&	\text{Inv-}\Gamma(\lambda_{\text{\tiny$\Delta$}}\,\boldsymbol{|}\,a'',b'')\\
\label{eq:sparse_model_summary}          \sigma_n^2 &\sim&	\text{Inv-}\Gamma(\lambda_{\text{\tiny$\Delta$}}\,\boldsymbol{|}\,a_\sigma,b_\sigma)\,.
\end{eqnarray}
We also define $\zeta\in\mathcal{C} = \{\kWeibull,\lWeibull,\lambda_{\text{\tiny$\Delta$}},\sigma_n^2\}$ as a representative variable with 
corresponding positive, real-valued parameters $a_\zeta\in\{a,a'',a_\sigma\}$ and $b_\zeta\in\{a,a'',a_\sigma\}$, that belong to the respective 
density functions. Further, the set $\mathcal{C}_{\setminus\zeta}$ denotes the set $\mathcal{C}$ without the respective variable $\zeta$.
Fig. \ref{fig:dependency_graph} shows a graphical model that helps to visualize the dependencies in this model. %
Herein, $\theta$ and $\boldsymbol{\Xi}$ are only valid for strategy $\textbf{\emph{S2}}$, which is discussed in \mbox{Section \ref{sec:PDL}.} 
For the particular model in (\ref{eq:sparse_model_summary}), we assume that the variables $\x, \sigma_n^2$ and $\theta$ are mutually independent.
Gibbs sampling requires the full conditional distributions for each parameter of interest. 
Based on these assumptions, we obtain the relation %
\begin{eqnarray}
	\p{\zeta}{\y,\x,\mathcal{C}_{\setminus\zeta}} & \propto & \p{\y}{\x,\mathcal{C}}\,\p{\zeta}{\x,\mathcal{C}_{\setminus\zeta}}\\
												  & \propto & \p{\y}{\x,\mathcal{C}}\, \p{\zeta}{\mathcal{C}_{\setminus\zeta}}\, \tilde{p}(\x\,\boldsymbol{|}\,\mathcal{C}) \,.
\end{eqnarray}
Since the prior distributions are all conjugate to the Gaussian likelihood function in (\ref{eq:Gauss_likelihood}), a simple calculation yields the posterior distributions of the parameters involved in the Gibbs sampling procedure. 
For $\zeta\in\mathcal{C}_{\setminus\kWeibull}$, we obtain  
\begin{equation}
\label{eq:posterior_zeta_no_kw}
	\zeta\,\boldsymbol{|}\,{\y,\x,\mathcal{C}_{\setminus\zeta}} \, \sim\ \text{Inv-}\Gamma(\zeta\,\boldsymbol{|}\,a_\zeta+{\text{\scriptsize$\frac{M}{2}$}},\ b_\zeta +\text{\scriptsize$\frac{1}{2}$}\,)\ \tilde{p}(\x\,\boldsymbol{|}\,\mathcal{C}), %
\end{equation}
and for $\kWeibull$, we obtain  %
\begin{equation}
\label{eq:posterior_kw}
	\kWeibull\,\boldsymbol{|}\,{\y,\x,\mathcal{C}_{\setminus\kWeibull}}\ \sim\ \ \p{\kWeibull}{\tilde{a}\,',\, \tilde{b}\,',\, \tilde{c}\,' }\ 
	\tilde{p}(\x\,\boldsymbol{|}\,\mathcal{C}),
\end{equation}
with parameters
$\tilde{a}\,' = a\,'+N$, $\ \tilde{b}\,'=b\,'+\sum_{i=1}^{N}\log(x_i)$, and $\tilde{c}\,' = (d\,')^{\kWeibull}+\sum_{i=1}^{N}x_i^{\kWeibull}$.
Samples of the posterior variables can be obtained using Metropolis Hastings \cite{Bishop2006} or HMC. \\  %

\noindent The sparse coefficients are sampled using HMC. 
We briefly describe the idea of this method according \mbox{to \cite{Neal2011},} adapted to our model for $\x$:\\ %
Within the framework of HMC, the sampling process is described in terms of \emph{Hamilton dynamics}, a concept known from classical physics. 
It is used to describe the 
trajectory of a physical system in phase space, based on its potential and kinetic energy. 
HMC assigns to every sparse coefficient, $x_i$, an associated momentum variable, $\xi_i$, $i=1,\dots,N$, that is responsible for 
the sampling dynamics. %
The posterior density to be sampled is related to the potential energy, given by \cite{Neal2011} 
\begin{equation}
	U(\x\,\boldsymbol{|}\,\y,\mathcal{C})\ =\ -\log \tilde{p}(\x\,\boldsymbol{|}\,\y,\mathcal{C}) - \log(Z_u)\,,
\end{equation}
where $Z_u$ is a suitable normalization constant. 
Since $\y$ and $\mathcal{C}$
are fixed, we may drop them and write $U(\x)$ instead. %
The kinetic energy, $K(\boldsymbol{\xi})$, depends only on the auxiliary variables $\boldsymbol{\xi}\!=\![\xi_1,\dots,\xi_N\!]$. 
A standard choice for $K(\boldsymbol{\xi})$ %
corresponds to independent particles in free space with mass $m_i$, i.e. \mbox{$K(\boldsymbol{\xi}) = \sum_{i=1}^{N}\xi_i^2/2m_i$.}   %
The dynamics of the sampling process are governed by the \emph{Hamiltonian function}, which is given by $\mathcal{H}(\x,\boldsymbol{\xi})\! =\!U(\x)\! +\! K(\boldsymbol{\xi})$ and
represents the total system energy. %
The %
joint density of  $\!(\x,\boldsymbol{\xi})\!$ is defined by \cite{Neal2011} %
\begin{equation}
\label{eq:HMC_canonical_density}
 	p(\x,\boldsymbol{\xi}) =  \frac{1}{Z_c} \text{e}^{-\frac{\mathcal{H}(\x,\boldsymbol{\xi})}{T_{\text{\scriptsize sys}}}}\!
                                   = \ \tilde{p}(\x\,\boldsymbol{|}\,\y,\mathcal{C})  \,\prod_{i=1}^{N}\mathcal{N}(\,\xi_i\,\boldsymbol{|}\,0,m_i\,).
\end{equation}
Herein, $T_{\text{\scriptsize sys}}$ is called the \emph{system temperature} and $Z_c$ is a normalization constant.
The last equation is obtained by setting $T_{\text{\scriptsize sys}}=1$ and $Z_u = Z_c$, 
while the Gaussian density arises from the special choice of the kinetic energy term.
In HMC, a proposal for a new sample is obtained by the final points ($x_i^*,\xi_i^*$)
of a trajectory described by Hamilton's equations of motion.
They are calculated $\forall\ (x_i,\xi_i), i\!=\!1,\dots,N$, \cite{Neal2011}:
\begin{equation}
\hspace{0.0cm}	\frac{\mathrm{d}x_i}{\mathrm{d}t} = \frac{\xi_i}{m_i}\,, \qquad \frac{\mathrm{d}\xi_i}{\mathrm{d}t} = -\frac{\partialfrac{}{x_i}\tilde{p}(\x\,\boldsymbol{|}\,\y,\mathcal{C}) }{\tilde{p}(\x\,\boldsymbol{|}\,\y,\mathcal{C}) }\,.
\end{equation}
A Metropolis update decides, whether a proposed sample is accepted or rejected, with %
acceptance probability \cite{Neal2011}
\begin{equation}
	\text{P}(\text{accept}) = \min_{}\,(\,1 ,\ \exp{-\mathcal{H}(x_i^*,\xi_i^*) + \mathcal{H}(x_i,\xi_i)}\ )\,.
\end{equation}

\section{Parametric DL strategies for CFS} 
\label{sec:PDL}
\begin{figure}
\centering
	\includegraphics[width=0.63\columnwidth]{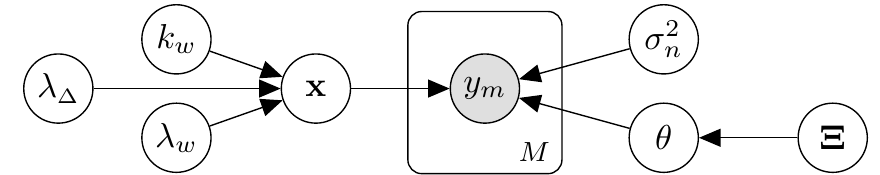}
\caption{Dependency relations for the complete hierarchical model. The variables $\theta$ and $\boldsymbol{\Xi}$ appear exclusively in \textbf{\emph{S2}}.}
\label{fig:dependency_graph}
\end{figure}

\noindent
In this section, we present two strategies for parametric dictionary learning in CFS. 
In the first \mbox{strategy (\textbf{\emph{S1}}),} we follow the ideas of hybrid Bayesian inference \cite{Yuan2009,Yuan2015}
and AM-based DL \cite{Beck2013}, %
where $\theta$ is a deterministic parameter, that is estimated using the Monte Carlo EM algorithm in \cite{Bishop2006}. 
In the second \mbox{strategy (\textbf{\emph{S2}}),} we pursue a full Bayesian approach and consider a probabilistic model for $\theta$.
Herein, approximate inference is accomplished by extending the Gibbs sampler in \mbox{Section\! \ref{sec:approx_inference} to} jointly estimate 
$(\x,\theta,\sigma_n^2)$.
\mbox{Fig.\! \ref{fig:dependency_graph}} depicts the dependency graph for both strategies, where $\theta, \boldsymbol{\Xi}$ belong exclusively \mbox{to \textbf{\emph{S2}}.}\\ %
As pointed out in \cite{Yuan2009,Yuan2015}, hybrid and full Bayesian strategies have their individual advantages in certain situations.
For small sample sizes, %
Bayesian methods can be superior if good prior knowledge is available \cite{Yuan2009}.
Nonetheless, they are often computationally more complex and insufficient prior information can lead to a
small-sample bias, even if a non-informative prior is used \cite{Yuan2009}. 
In CFS, the sample size is small and only vague prior knowledge of $\theta$ is available. 
Therefore, we investigate the performance of both DL strategies based on our probabilistic sparse model. 
The computational complexity of both strategies is comparable. It is dominated by HMC, i.e. by sampling the high-dimensional vector $\x$ in each iteration of the Gibbs sampler.
Regarding $\theta$, the following prior knowledge is assumed:
In \textbf{\emph{S1}}, we roughly restrict the range of values \mbox{that $\theta$} can take, while in \textbf{\emph{S2}}, we define a non-informative prior
over the same range.
Recall that $\theta$ effectively describes the filter characteristics of the lowpass \mbox{filter $H_{\text{\tiny LP}}(\omega)$.}
To create the dictionary for a certain value of $\theta$ using (\ref{eq:dict_atoms_elements}), the inverse Fourier transform in 
(\ref{eq:sensor_signa_IFT_model}) has to be 
evaluated for each atom. Thus, the dictionary is not a simple function of $\theta$ and we restrict ourselves to a discrete set of 
parameters, with lower and upper bound, $\theta_{\text{\tiny min}}$ and $\theta_{\text{\tiny max}}$, respectively.
Since the bandwidth should be positive and bounded, we have $0 < \theta_{\text{\tiny min}}$ and $\theta_{\text{\tiny max}} < \infty$.
Then, the set $\Theta$ contains the discrete values $\theta_r, r=1,\dots,\Rtheta$,

\subsection{\hspace{-0.0cm}Hybrid DL: iterative estimation of $\theta$ and $(\x,\mathcal{C})$  (\textbf{\emph{S1}})\hspace{-0.0cm}}
\noindent
The dictionary parameters in the CFS problem can be iteratively estimated using a Monte Carlo EM algorithm.
First, an initial value, $\theta^{(0)}\!$, has to be chosen.
In subsequent iterations with indices $d\!=\!1,\dots,d_{\text{\tiny max}}$, 
we obtain joint samples $\{\x_l,\mathcal{C}_l\}^{(d)},\,l=1,\dots,L_{\text{\tiny MC}}$, %
by Gibbs sampling and HMC according to \mbox{Section \ref{sec:approx_inference}.} 
Then, we determine the posterior expectation 
of $\zeta\in \mathcal{C}$, using the previous estimate $\hat{\theta}^{(d-1)}$:
\begin{eqnarray}
	\hat{\zeta}^{(d)}  &=& \int_{\text{dom}(\zeta)} \zeta \,\p{\zeta}{\y,\hat{\theta}^{(d-1)}}\,\mathrm{d}\zeta \\
\label{eq:posterior_mean_zeta} 	&\approx&\!\! \frac{1}{L_{\text{\tiny MC}}}\sum_{l=1}^{L_{\text{\tiny MC}}}\ \zeta_{\,l}^{(d)}\, \p{\zeta_l^{(d)}}{\y,\hat{\theta}^{(d-1)}},
\end{eqnarray} 
where $\text{dom}(\zeta)$ is the domain of $\zeta$. 
The current estimates of the reflection delays, $\hat{\mathcal{S}}^{(d)}$, are determined by identifying the indices of the $K$ largest 
elements in the posterior mean of $\x$, denoted by $\hat{\x}^{(d)}$. It is obtained by exchanging $\zeta_{\,l}^{(d-1)}$\!\! with $\x_l^{(d-1)}\!\!$ in (\ref{eq:posterior_mean_zeta}). %
Besides, we also estimate the amplitudes of the significant components in $\x$. 
They can be useful to assess the sparsity level 
of the solution and to determine the amount of optical power reflected from the FBGs. 
Since the posterior of $\x$ is multimodal with one narrow peak around zero and another peak at some larger amplitude,
the MAP is more suitable for this task. %
It is given by %
\begin{eqnarray}
\label{eq:EM_MAP_x}
 \hspace{-0.5cm}	\{\mathbf{\hat{x}},\hat{\mathcal{C}}\,\}_{\text{\tiny MAP}}^{(d)}\hspace{-0.1cm} &=&\hspace{0.05cm}  \text{arg}\max_{\hspace{-0.25cm}\x,\mathcal{C}}\ \log \p{\x,\mathcal{C}}{\y,\hat{\theta}^{(d-1)}}\\[0.2cm]%
\label{eq:EM_MAP_x_approx}	     \hspace{-0.5cm}    &\hspace{-2.8cm}\approx&\hspace{-0.6cm} \text{arg}\hspace{-1.08cm}\max_{\hspace{-0.45cm}\{\x_j,\mathcal{C}_j\}\in\{\x_l,\,\mathcal{C}_l\}^{(d)}_{l=1,..,L_{\text{\tiny MC}}}}\hspace{-0.5cm}  \log \p{\{\x_j,\mathcal{C}_j\}^{(d)}}{\y,\hat{\theta}^{(d-1)}}.
\end{eqnarray}
However, the estimates of $\mathcal{S}$ obtained from $\hat{\x}_{\text{\tiny MAP}}^{(d)}$ are less accurate than those obtained by the posterior mean.
Therefore, the empirical MAP solution is only used to estimate the reflection amplitudes.
Next, we calculate the current estimate $\hat{\theta}^{(d)}$ by taking the expected value over $\x,\mathcal{C}$ given 
$\y,\theta$ (E-step): %
\begin{eqnarray}
\label{eq:EM_E_step}
\nonumber\hspace{-1.0cm}	 && \!\!\! \EXPC{\x,\!\mathcal{C}\,}{\,\y,\!\theta}{} \log \p{\y,\x,\mathcal{C}}{\theta}\\  %
\hspace{-1.0cm}				&=&\!\!\!\! \int_{\mathbb{R}_+^N} \int_{\Psi}  
						    \log \p{\y,\x,\mathcal{C}}{\theta}\, \p{\x,\mathcal{C}}{\y,\theta}\,\mathrm{d}\mathcal{C}\,\mathrm{d}\x\\ %
\label{eq:Qfct}\hspace{-1.0cm} 			    &\approx&\!\!\!\! \frac{1}{L_{\text{\tiny MC}}}\sum_{l=1}^{L_{\text{\tiny MC}}}\ \log \p{\y,\{\x_l,\mathcal{C}_l\}^{(d-1)}\!}{\theta}\  \triangleq\ Q(\,\theta\,\boldsymbol{|}\,\hat{\theta}^{(d-1)}).
\end{eqnarray}
Herein, $\Psi$ is the product space formed by the individual domains of all variables in $\mathcal{C}$. 
In the $M$-step, a locally optimal value, $\hat{\theta}^{(d)}$, is obtained by maximizing $\theta$ over the set $\Theta$, i.e.
\begin{equation}
\label{eq:EM_M_step}
	\hat{\theta}^{(d)}\  =\  \text{arg}\max_{\hspace{-0.3cm}\theta\, \in\, \Theta}\ Q(\,\theta\,\boldsymbol{|}\,\theta^{(d-1)}\,)\,.
\end{equation}
\vspace{0.1cm}
\subsubsection{Initialization of $\theta$ via bisectional search} 
An adequate initialization, $\theta^{(0)}\!$, can alleviate the problem of local optima in the EM algorithm. 
In CFS, the desired sparsity level is known to be the number of reflections, $K$. 
Hence, a good choice for $\theta^{(0)}\!$ yields a solution for $\x$ with $K$ significant non-zero elements.
Starting at an arbitrary value $\theta^{(0)}\!\in\Theta$, a bisectional search within $\Theta$ can quickly determine a suitable initial value.
After choosing the first value at random, $\Theta$ is subdivided into two parts, containing all larger and all smaller values, respectively. %
When the number of peaks is too high, the next trial is chosen as the median of the lower division. If it is too low, the next trial is the median
of the upper division, and so on.
For a properly selected $\theta^{(0)}\!$, \textbf{\emph{S1}} converges faster and is more likely to approach (or even attain) the global optimum.

\subsection{Bayesian DL: joint estimation of ($\x,\mathcal{C},\theta$)$\ $ (\textbf{\emph{S2}})}
\noindent 
In strategy \textbf{\emph{S2}}, we treat $\theta$ as a random variable. %
Due to its discrete nature, each element \mbox{$\theta_r\in\Theta$} is assigned a probability mass, $p_r = p(\theta_r)\,, r=1,\dots,\Rtheta$,
where $\sum_{r=1}p(\theta_r)=1$. 
Then, $\theta$ is \mbox{categorically (Cat)} distributed over the set of discrete dictionary parameters, $\Theta$, with corresponding probability masses in
$\boldsymbol{\Xi} = \{p_1,\dots,p_{\text{\tiny $R_{\small\Theta}$}}\}$. 
Uncertainty in the \emph{a priori} assigned probability masses is taken into account in terms of a prior on $\boldsymbol{\Xi}$.
The Dirichlet (Dir) distribution can be used as the conjugate prior %
with parameters $\boldsymbol{\nu} = [\nu_1,\dots,\nu_{\text{\tiny$\Rtheta$}}]^{\!\top}$, i.e.
\begin{equation}
	p(\boldsymbol{\Xi})	=\ \frac{1}{B(\boldsymbol{\nu})}\prod_{r=1}^{\Rtheta}p_r^{\nu_r}\,,
\end{equation}
where $B(\boldsymbol{\nu})$ denotes the \emph{Beta} function and the variables $\nu_r$, $r=1,\dots,\Rtheta$, describe the number of occurrences of the values in $\Theta$. %
When a new element, $\theta_q\in\Theta$, is sampled, a posterior count is assigned to that value. 
After sampling another value in the next iteration, this count is reassigned to the new value.
Let $\mathbf{\breve{c}}\in \mathbb{N}^{\Rtheta}$ indicate the current sample, i.e. $c_q=1$ for one index  $q\in\{1,\dots,\Rtheta\}$,  %
while all other elements are zero.
A non-informative prior is obtained if all values $\theta_q\in\Theta$ are equally likely and each element is assigned a single count.
Then, $\nu_r=1\ \forall\ r=1,\dots,\Rtheta$ and a new sample has a strong impact on the posterior distribution. 
In contrast, for large values, e.g. $\nu_r\! =\! 1000\ \forall\ r\!=\!1,\dots,\Rtheta$, 
a new count leaves the distribution \mbox{almost invariant.}   %
The complete model is then given by (\ref{eq:sparse_model_summary}) and, in addition,
\begin{eqnarray}
	\boldsymbol{\Xi}  &\sim & \text{Dir}(\,\boldsymbol{\Xi}\,\boldsymbol{|}\,\boldsymbol{\nu}\,)\\[0.15cm]
	\theta\,\boldsymbol{|}\,\boldsymbol{\Xi} &\sim& \text{Cat}(\theta\,\boldsymbol{|}\,\Rtheta,\boldsymbol{\Xi})  \, .
\end{eqnarray}
To accomplish approximate inference in this model, the variables $\theta$ and $\boldsymbol{\Xi}$ are included in the Gibbs sampling 
procedure of Section \ref{sec:approx_inference}. Therefore, the conditional distributions must be determined.
Based on the dependencies in Fig. \ref{fig:dependency_graph}, and since $\x,\sigma_n^2$ and $\theta$ are assumed to be mutually independent, we find 
\begin{eqnarray} %
\label{eq:theta_Gibbs}
	\boldsymbol{\Xi} \,\boldsymbol{|}\,\theta\ =\ \boldsymbol{\widetilde{\Xi}}\ & \sim &\ \text{Dir}(\,\boldsymbol{\Xi}\,\boldsymbol{|}\,\boldsymbol{\nu} + \mathbf{\breve{c}}\,),\\[0.15cm]
\label{eq:Xi_Gibbs}
	\theta \,\boldsymbol{|}\,\y,\boldsymbol{\widetilde{\Xi}}\ &\sim&\ \text{Cat}(\,\theta\,\boldsymbol{|}\,\Rtheta,\boldsymbol{\widetilde{\Xi}}\,).
\end{eqnarray}
\begin{figure}[t]
\begin{center}
\line(1,0){280}\\ 
\textbf{Algorithm:}\ Sparse estimation and PDL, strategy \textbf{\emph{S1}} \& \textbf{\emph{S2}}\vspace{-0.48cm}
\begin{center}
\line(1,0){280}
\end{center}\vspace{-0.05cm}
\begin{tabular}{ll}

\hspace{-0.1cm}\textbf{Input:}		& \hspace{-1.6cm}	  $\mathbf{y},M,\PHI,N,L,T_d,\delta t,r(t,\theta),K,L_{\text{\tiny MC}},d_{\text{max}}$\\
\hspace{-0.1cm}\textbf{Output:}	& \hspace{-1.6cm}	  $\hat{\mathcal{S}},\hat{\mathbf{x}},\hat{\theta},\hat{\sigma}_n,d,ee  $\\
\hspace{-0.1cm}\textbf{Parameters:}& \hspace{-1.6cm}	  $a,a',a'',a_\sigma,b,b',b'',b_\sigma,d\,',\boldsymbol{\nu}, \Rtheta, \{\theta_r\}_{r=1}^{\Rtheta}$,\\
				    & \hspace{-1.6cm} 	  internal HMC parameters (c.f. \cite{Neal2011,Homan2014}).\\
\hspace{-0.1cm}\textbf{0.\, Initialize:}& \hspace{-1.55cm} 	  $\theta$ at random $\rightarrow  \hat{\theta}^{(0)}$ via bisectional search,\\
					& \hspace{-1.55cm} 	  $\mathbf{A}(\hat{\theta}^{(0)}),\{\hat{\mathbf{x}}^{(0)}\!\!,\, \hat{\mathcal{C}}^{(0)}\}\!$ as in (\ref{eq:sparse_model_summary}), (\textbf{\emph{S2}}):$\,d_{\text{max}}\!\!=\!1$\\
\hspace{-0.1cm}\textbf{1.\, for} $d = 1$ to $d_{\text{max}}$ \textbf{do} \hspace{-0.6cm}	& \\
\hspace{-0.1cm}\textbf{2.} 		& \hspace{-2.7cm} \textbf{for} $l=1$ to $L_{\text{\tiny MC}}$ \textbf{do}\\
\hspace{-0.1cm}\textbf{3.} 		& \hspace{-2.3cm} Gibbs sampling: (i) $\mathcal{C}_l^{(d)}$ using (\ref{eq:posterior_zeta_no_kw}) and (\ref{eq:posterior_kw}),\\
					& \hspace{0.1cm} 				 (ii) $\x_l^{(d)}$\! via HMC.\\
					& \hspace{-0.89cm} 		(\textbf{\emph{S2}}):$\,$ (iii) $\theta_l^{(d)}\!\!,\, \boldsymbol{\Xi}_l^{(d)}$\!\! using (\ref{eq:theta_Gibbs}) and (\ref{eq:Xi_Gibbs})\\
\hspace{-0.1cm}\textbf{4.} 		& \hspace{-2.7cm} \textbf{end for} \\
\hspace{-0.1cm}\textbf{5.} 		& \hspace{-2.7cm} Estimate:\hspace{0.28cm} $\hat{\mathcal{S}}^{(d)}$ from $\hat{\x}^{(d)}$ in (\ref{eq:posterior_mean_zeta}) with $\zeta_{\,l}^{(d)}\!\rightarrow \x_l^{(d)}\!\!$,\\
	& \hspace{-0.95cm} 					      					  $\hat{\mathcal{C}}^{(d)}$ from (\ref{eq:posterior_mean_zeta}),
											   \hspace{0.2cm} $\hat{\x}^{(d)}_{\text{\tiny MAP}}$ from (\ref{eq:EM_MAP_x_approx}),\\				
\hspace{-0.1cm}\textbf{5.a} 		& \hspace{-1.82cm}  (\textbf{\emph{S1}:}) $\,\hat{\theta}^{(d)} = \text{arg}\max_{\theta\in\Theta}\ Q(\theta\,\boldsymbol{|}\,\hat{\theta}^{(d-1)})$.\\
\hspace{-0.1cm}\textbf{5.b} 		& \hspace{-1.82cm}  (\textbf{\emph{S2}:}) $\,\hat{\theta}^{(d)}$ from (\ref{eq:posterior_mean_zeta}) with $\zeta_{\,l}^{(d)}\!\rightarrow \theta_l^{(d)}$.\\
\hspace{-0.1cm}\textbf{6.} 		& \hspace{-2.7cm}  \textbf{if}\ \ $\hat{\theta}^{(d)} ==\, \hat{\theta}^{(d-1)}$\ \ \ \textbf{or}\ \ \ $d==d_{\text{max}}$\\
\hspace{-0.1cm}\textbf{7.} 		& \hspace{-2.35cm}  \textbf{return} $\, \hat{\mathcal{S}}^{(d)}\!,\, \hat{\x}^{(d)}_{\text{\tiny MAP}},\,\hat{\mathcal{C}}^{(d)}\!,\, \hat{\theta}^{(d)}, ee\!=\!\|\y\!-\!\PHI\mathbf{A}(\hat{\theta}^{(d)}\!)\|_2^2$.\\
\hspace{-0.1cm}\textbf{8.} 		& \hspace{-2.7cm}  \textbf{end if}\\
\hspace{-0.1cm}\textbf{9. end for}
\end{tabular}\vspace{0.13cm}
\line(1,0){280}
\end{center}
\vspace{-0.3cm}
\end{figure}
\section{Simulations and experimental data}
\label{sec:performance}
 \noindent
Let us now evaluate the proposed sparse model and DL strategies. First, we show the qualitative behavior of the algorithms, 
followed by a quantitative performance analysis in comparison to the method in \cite{Weiss2016}. 
To this end, we consider several scenarios of different SNRs, CS sampling matrices and sample sizes.
Finally, we apply our algorithms to experimental data taken from a real fiber-optic sensor. 

\subsection{Simulation setup}
\noindent
We consider $K=3$ uniform FBGs in the sensing fiber, where the observed reflections have a common amplitude, $A_x$, and two reflections 
are closely spaced. Their delays are indicated by the inidces of the $K$ most significant elements in $\x$, contained in the set $\mathcal{S}$. 
Subsequently, the dictionary parameter is re-defined relative to its true value, i.e. $\hat{\theta}$ is replaced by $\hat{\theta}/\theta$.
Further, we use $\Rtheta = 100$ discrete parameter values, equally spaced between $30\%$ and $150\%$ of the true value.  %
The original signal (prior to CS) contains $L=134$ samples of the measured photocurrent. 
The dictionary atoms are created using $L$ samples of $r(t\!-\!i\delta t), i=1,\dots,N$, with a delay spacing of 
\mbox{$\delta t\!=\! 50$\! ns.} %
We use two types of CS matrices, $\PHI$, with i.i.d. entries drawn from the distributions below:
\begin{center}
\begin{tabular}{ll}
	(a) Gauss: 						& $\,\mathcal{N}(0,1)$,	\\
	(b) DF \cite{Achlioptas2003}	& $\{-1,0,1\}$ with probabilities $\{\frac{1}{6},\frac{2}{3},\frac{1}{6}\}$.
\end{tabular}
\end{center}
The variables $\{\mathcal{C},\theta,\boldsymbol{\Xi}\}$ are sampled according to \mbox{Section \ref{sec:approx_inference}.} 
For $\x$ we use the 'No-U-Turn' variant of the HMC \mbox{algorithm \cite{Homan2014},} which is efficiently implemented in 
the software package \emph{Stan} \cite{STAN}.
The algorithm \textbf{\emph{S1}} is initialized based on a bisectional search and runs at most $d_{\text{max}}=35$ iterations.
In \textbf{\emph{S2}}, we use a non-informative prior for $\theta$, with $p(\theta_r)=1/\Rtheta$ and %
$\nu_r = 1\ \forall\ r=1,\dots,\Rtheta$.  %

\subsection{Visualization and Working Principle}
\begin{figure}
\centering
  \subfigure[]{\includegraphics[width=0.34\columnwidth]{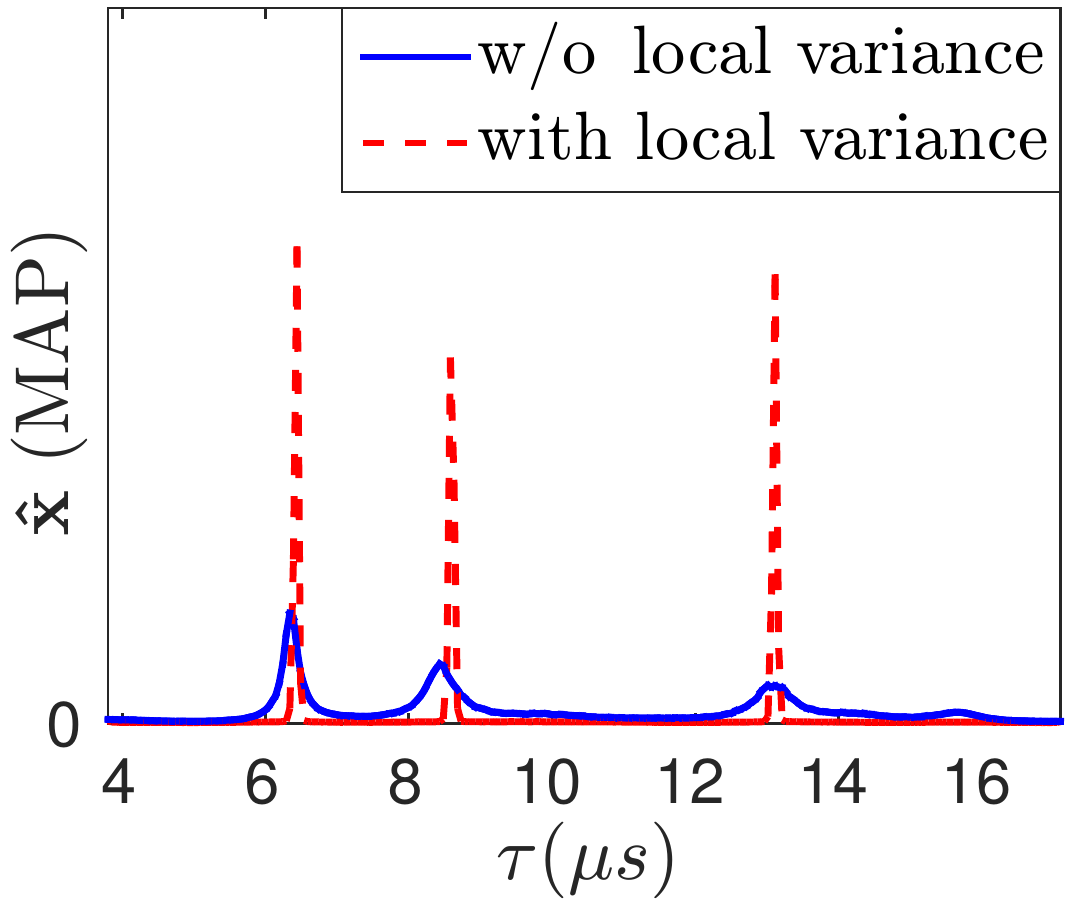}}\hspace{1.3cm}
  \subfigure[]{\includegraphics[width=0.34\columnwidth]{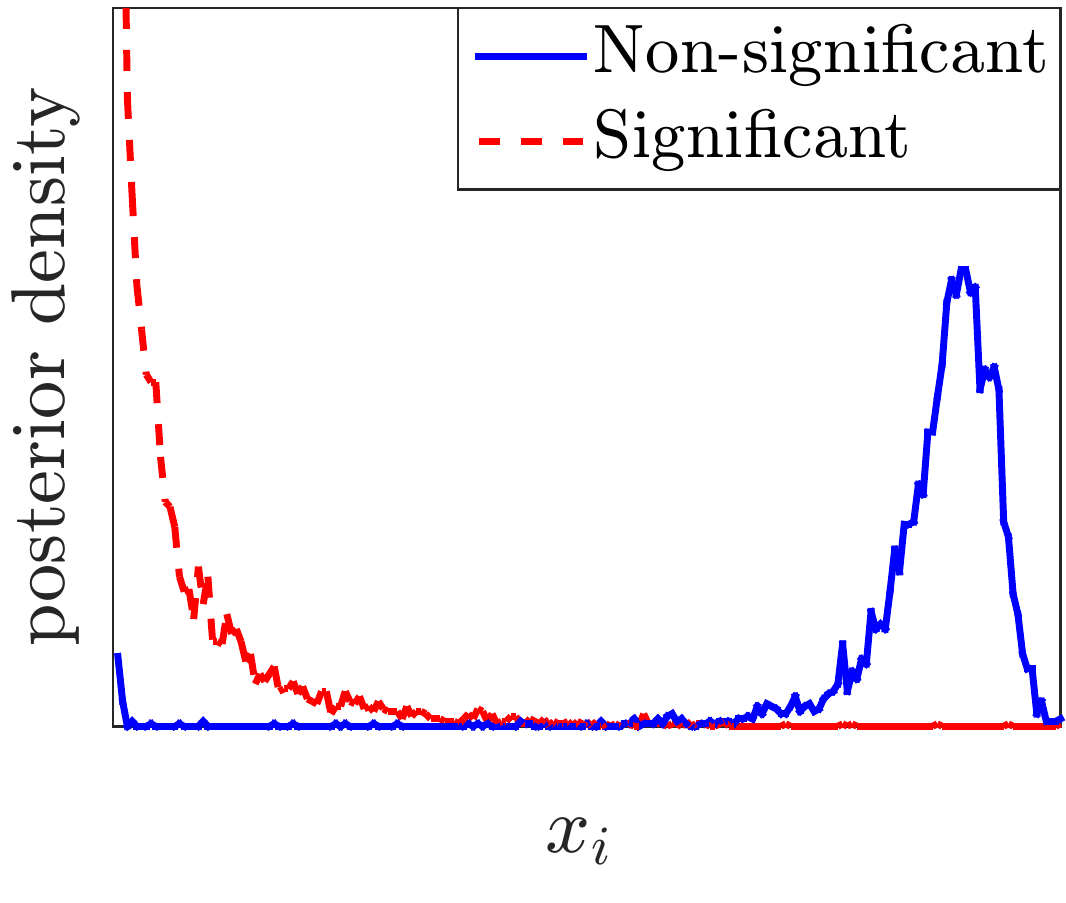}}\\
  \subfigure[]{\includegraphics[width=0.34\columnwidth]{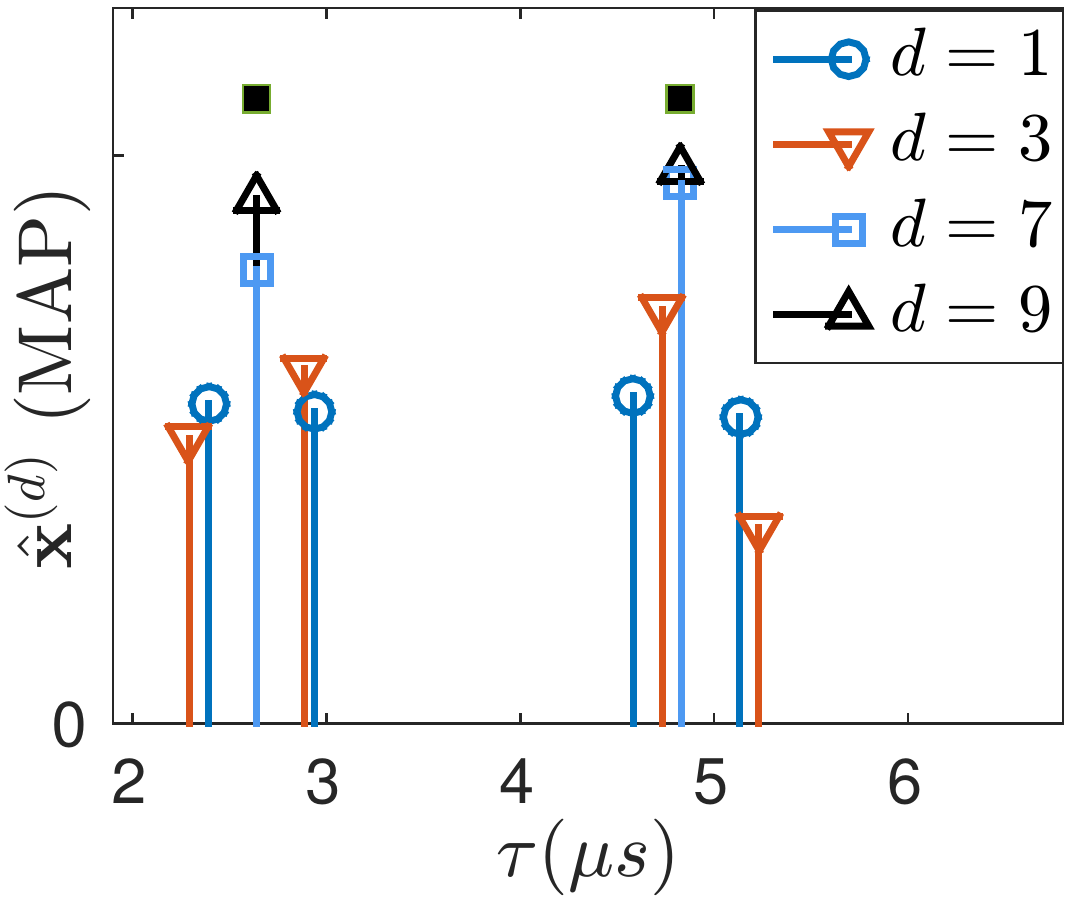}}\hspace{1.3cm}
  \subfigure[]{\includegraphics[width=0.34\columnwidth]{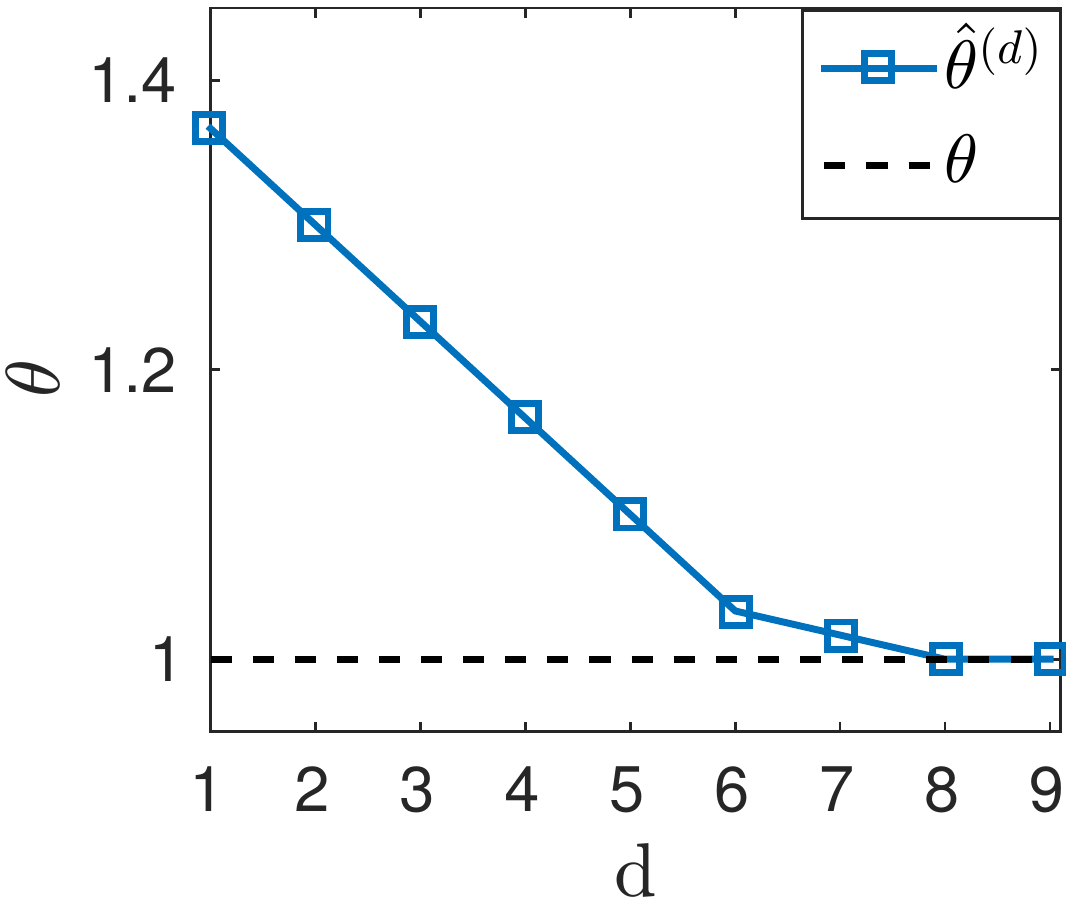}}\\
  \subfigure[]{\includegraphics[width=0.34\columnwidth]{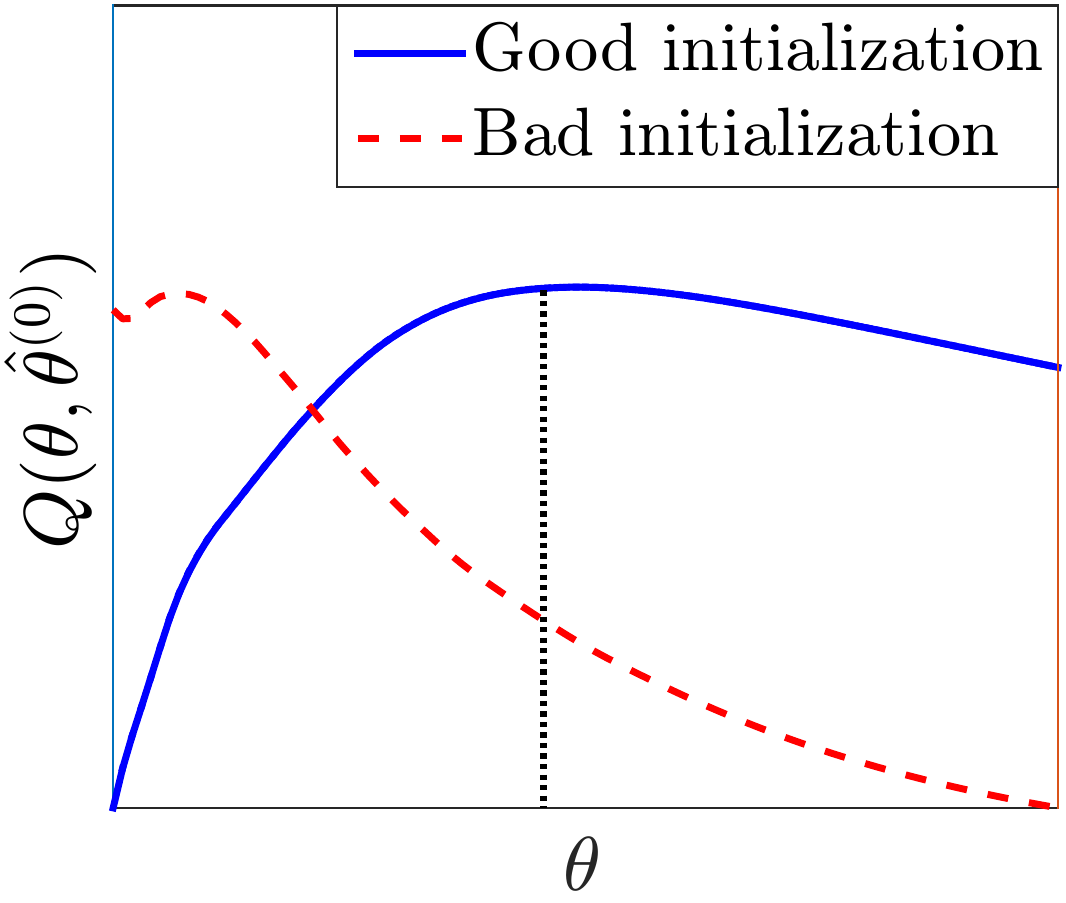}}\hspace{1.3cm}
  \subfigure[]{\includegraphics[width=0.34\columnwidth]{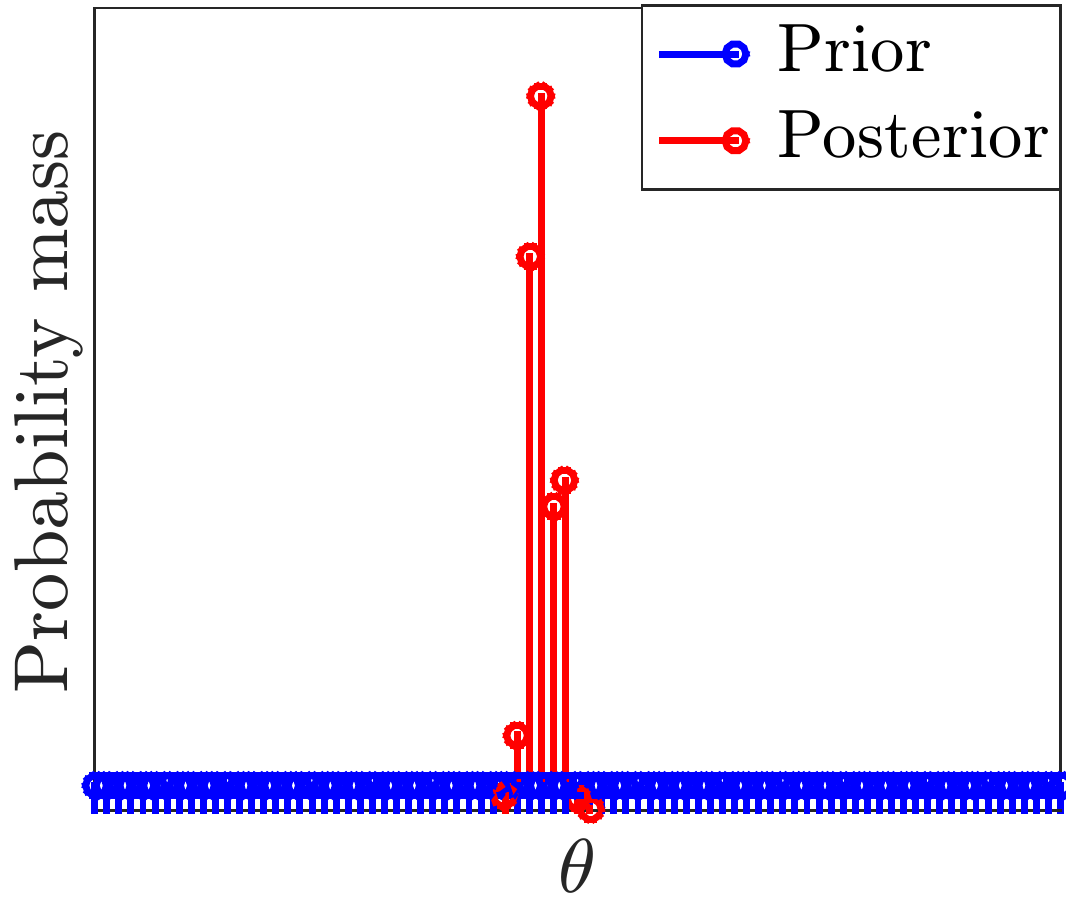}}
\caption{Visualization. Top left: empirical MAP solution for $\x$ and impact of collective shrinkage. 
		 Top right: empirical posterior density of a \mbox{non-/significant} entry in $\x$. Center left: evolution of the temporal MAP solution for $\x$ 
		 obtained by \textbf{\emph{S1}}. 
		 Center right: temporal solution for $\theta$ after the M-step. %
		 Bottom left: shape of the $Q$-function in \textbf{\emph{S1}} for a good and bad initial value $\theta^{(0)}$. 
		 Bottom right: for \textbf{\emph{S2}}, prior and posterior distribution of $\theta$ for $p(\theta_r)\!\!=\!\!1/\Rtheta$ and $\nu_r\! =\! 1,\,r\!=\!1,\dots,\!\Rtheta$.}\vspace{-0.2cm}
\label{fig:visualization}
\end{figure}

\noindent
The working principle of the algorithms is presented for \mbox{$\text{SNR} = 20$} dB and  
a Gaussian CS matrix using $M/L=50\%$ of the original samples.  %
Fig. \ref{fig:visualization} (top left) depicts the MAP solution for $\x$, obtained by HMC within Gibbs sampling according to 
Section \ref{sec:approx_inference}, where $\theta$ is fixed to the true value. 
It shows, that collective shrinkage, imposed by the local similarity assumption in the joint prior density of $\x$, yields a highly 
improved sparsity level in the presence of strong dictionary coherence.  
Fig. \ref{fig:visualization} (top right) shows the posterior density of $\x$ in one dimension. For a non-significant component, it 
is strongly peaked around zero, and for a significant component, is multimodal with a strong mode around the true amplitude and a 
smaller mode around zero. 
The second row in \mbox{Fig. \ref{fig:visualization}} delineates the evolution of the EM algorithm in \textbf{\emph{S1}} over several iterations. 
Fig. \ref{fig:visualization} (center left) shows the current MAP solutions for $\x$, i.e. $\x_{\text{\tiny MAP}}^{(d)}$, 
zoomed on
the two left-sided peaks. %
Due to a bad initial value for $\theta$, more than $K$ peaks appear in the first iterations. 
However, as the algorithm proceeds, significant peaks are formed only at the positions of the true significant components (black bullets). 
Fig. \ref{fig:visualization} (center right) shows, that also $\theta$ approaches the true value. 
Fig. \ref{fig:visualization} (bottom left) delineates a typical shape of the function $Q(\theta\,|\,\hat{\theta}^{(d-1)})$ of \textbf{\emph{S1}} in (\ref{eq:Qfct}), for a properly and
badly chosen initial value, $\theta^{(0)}$. A good choice leads to faster convergence, while for a bad choice, the algorithm might get either stuck 
at a local optimum or requires many EM iterations before the maximum of the Q-function appears close the true value of $\theta$.
Finally, Fig. \ref{fig:visualization} (bottom right) depicts for \textbf{\emph{S2}}, 
the non-informative prior of $\theta$ and a typical posterior density when $\nu_r=1\ \forall\ r=1,\dots,\Rtheta$. 

\subsection{Performance evaluation}  
\noindent
The performance is evaluated in terms of the root mean-squared error (RMSE). For a vector $\mathbf{v}$ and an estimator $\mathbf{\hat{v}}$, 
it is given by $\text{RMSE}(\mathbf{v},\mathbf{\hat{v}}) = (\EXP{}{\!\|\mathbf{v}-\mathbf{\hat{v}}\|_2^2})^{1/2}$.
\begin{figure} 
\centering
    \subfigure[]{\includegraphics[width=0.75\columnwidth]{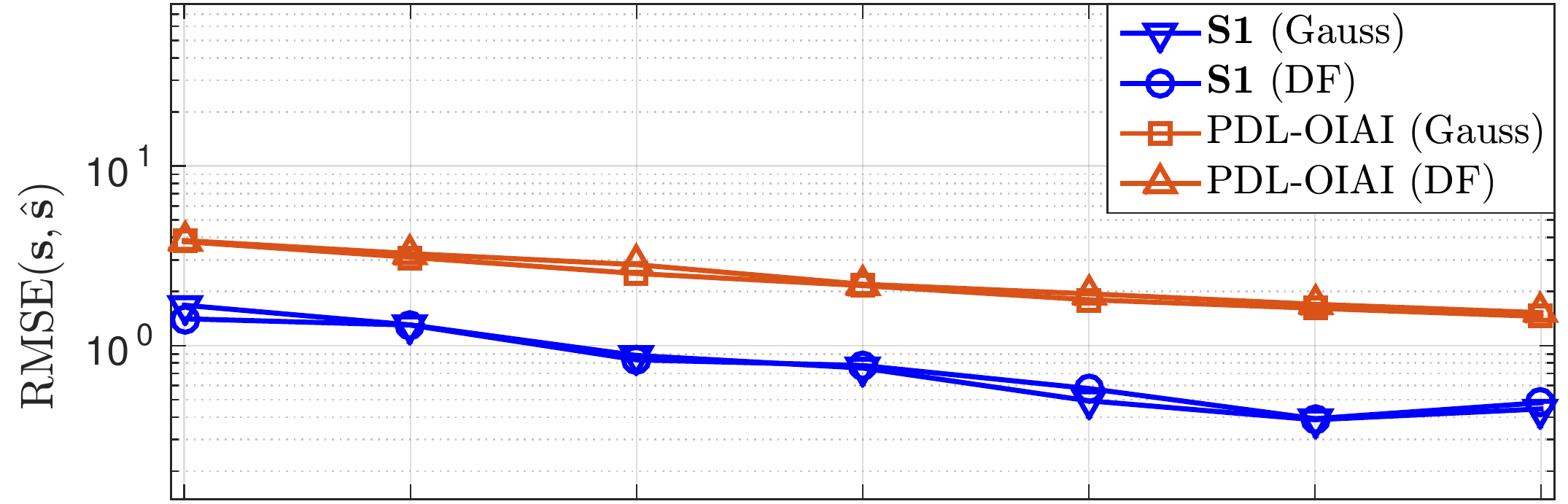}}\\[0.2cm]
    \subfigure[]{\includegraphics[width=0.75\columnwidth]{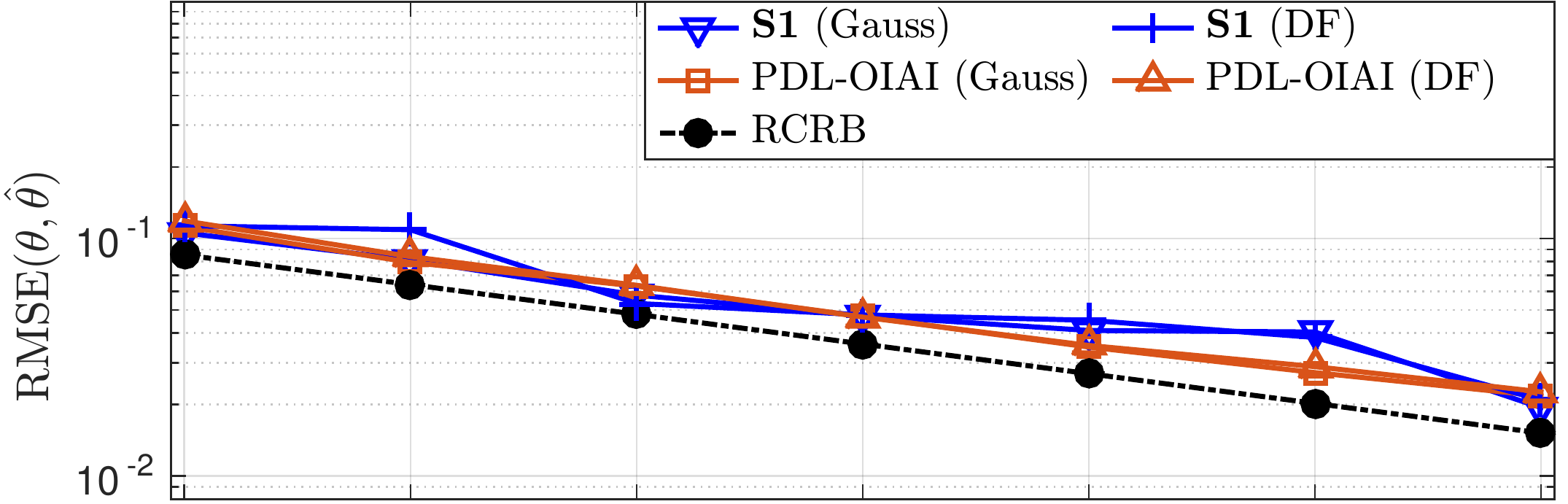}}\\[0.2cm]
    \subfigure[]{\includegraphics[width=0.75\columnwidth]{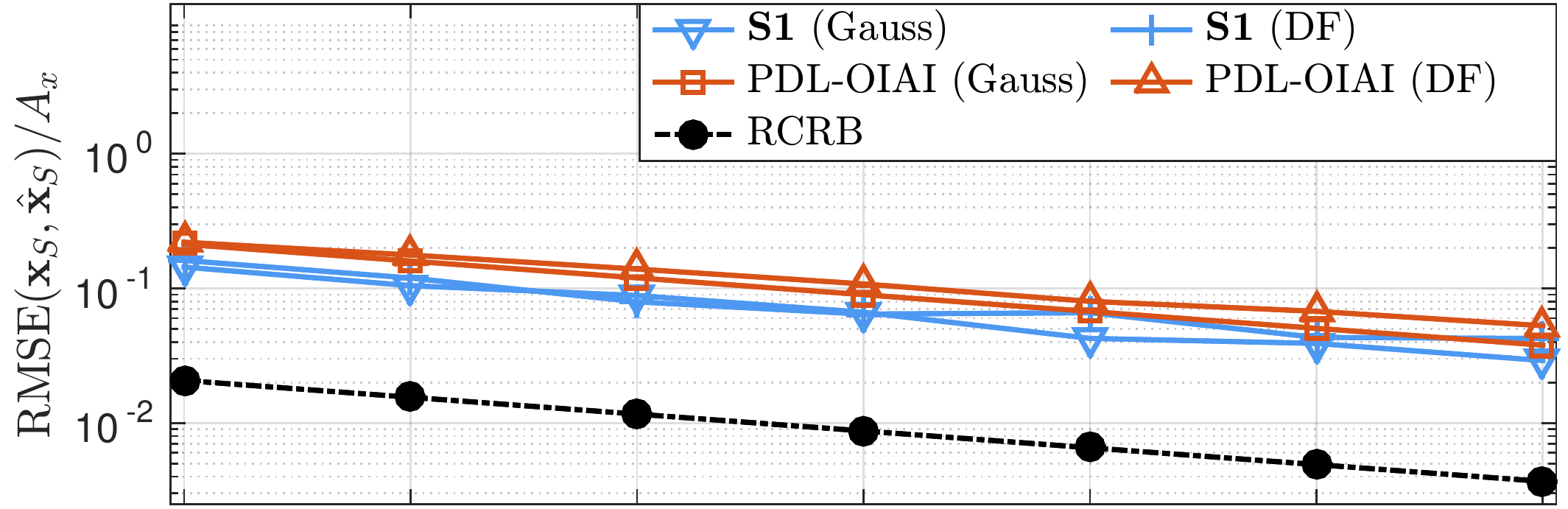}}\\[0.2cm]
    \subfigure[]{\includegraphics[width=0.75\columnwidth]{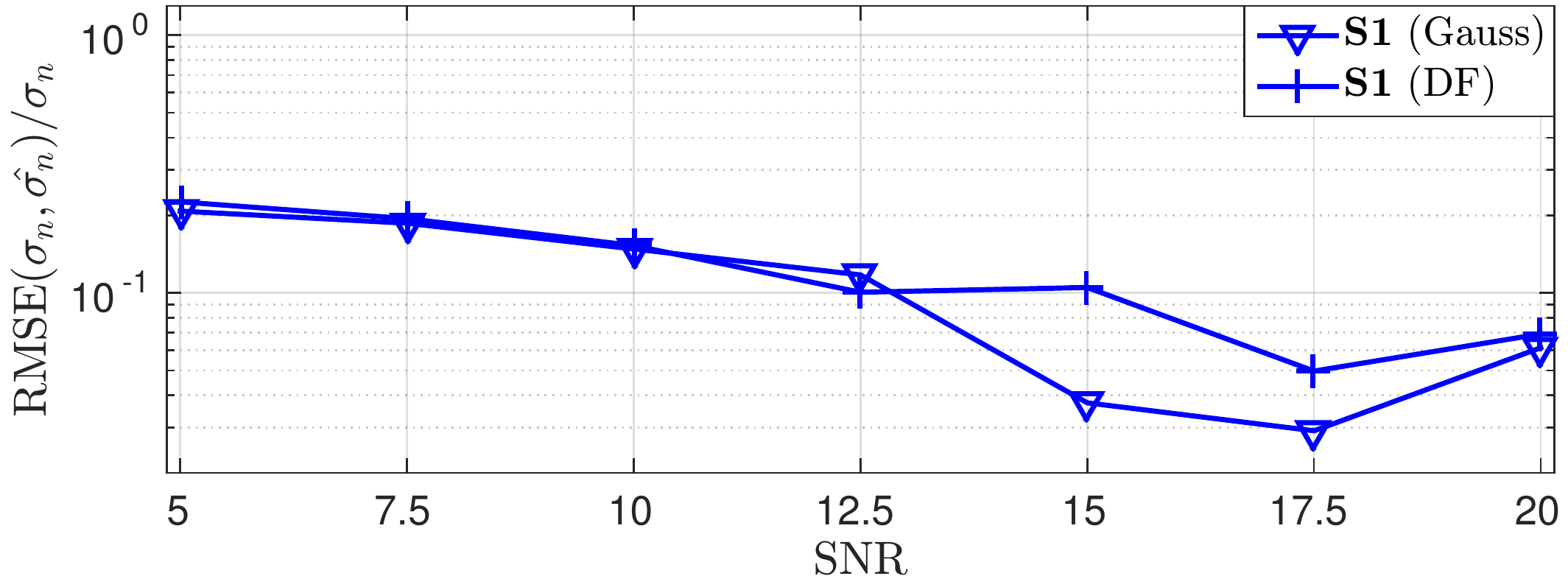}}
\caption{Performance of \textbf{\emph{S1}} in terms of the RMSE using $M/L=50\%$ of the original samples in comparison with PDL-OIAI in \cite{Weiss2016} and with the lower bound of the RMSE imposed by the CRB (RCRB). }
\label{fig:EM_performance_50}
\end{figure}
\begin{figure} 
\centering
    \subfigure[]{\includegraphics[width=0.75\columnwidth]{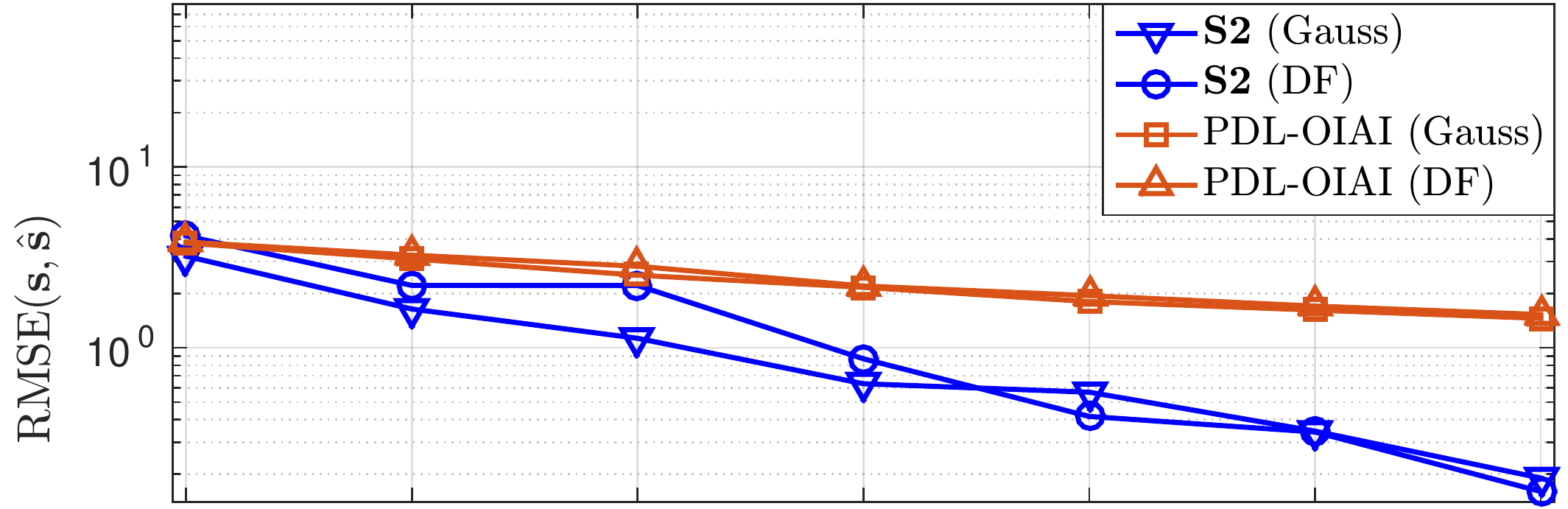}}\\[0.2cm]
    \subfigure[]{\includegraphics[width=0.75\columnwidth]{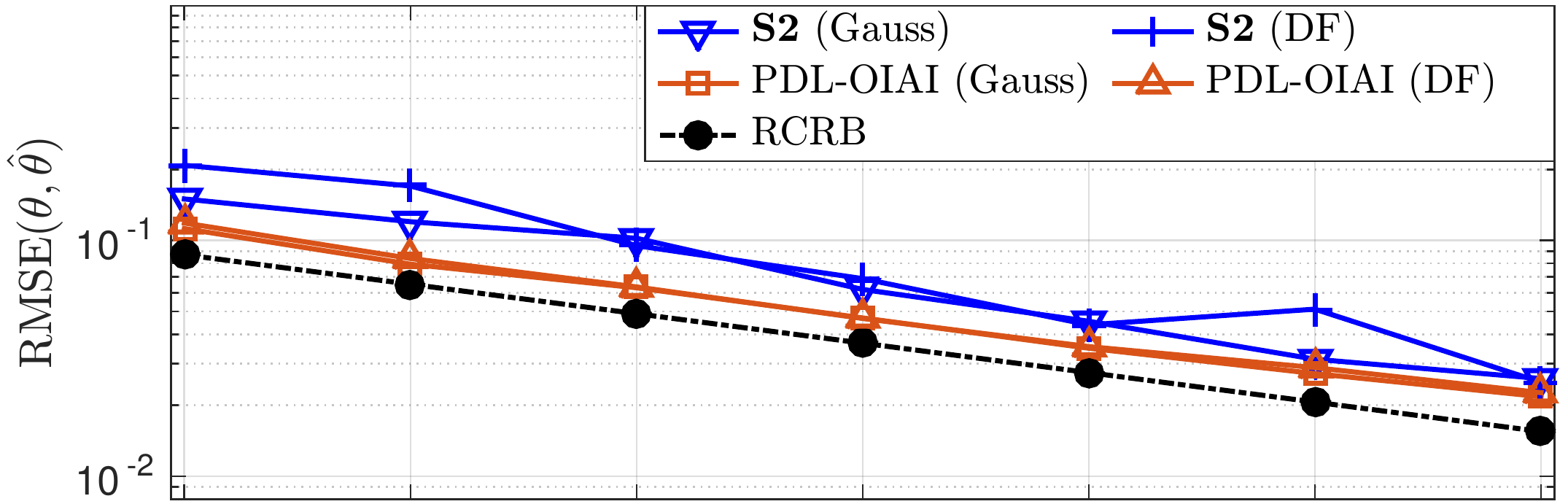}}\\[0.2cm]
    \subfigure[]{\includegraphics[width=0.75\columnwidth]{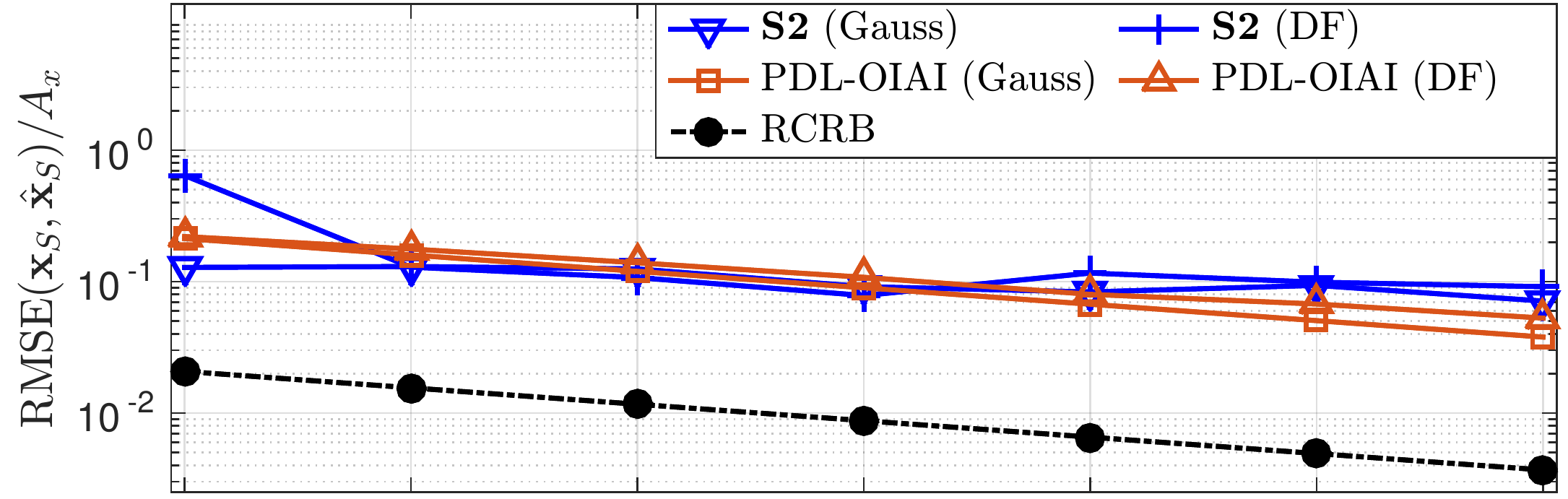}}\\[0.2cm]
    \subfigure[]{\includegraphics[width=0.75\columnwidth]{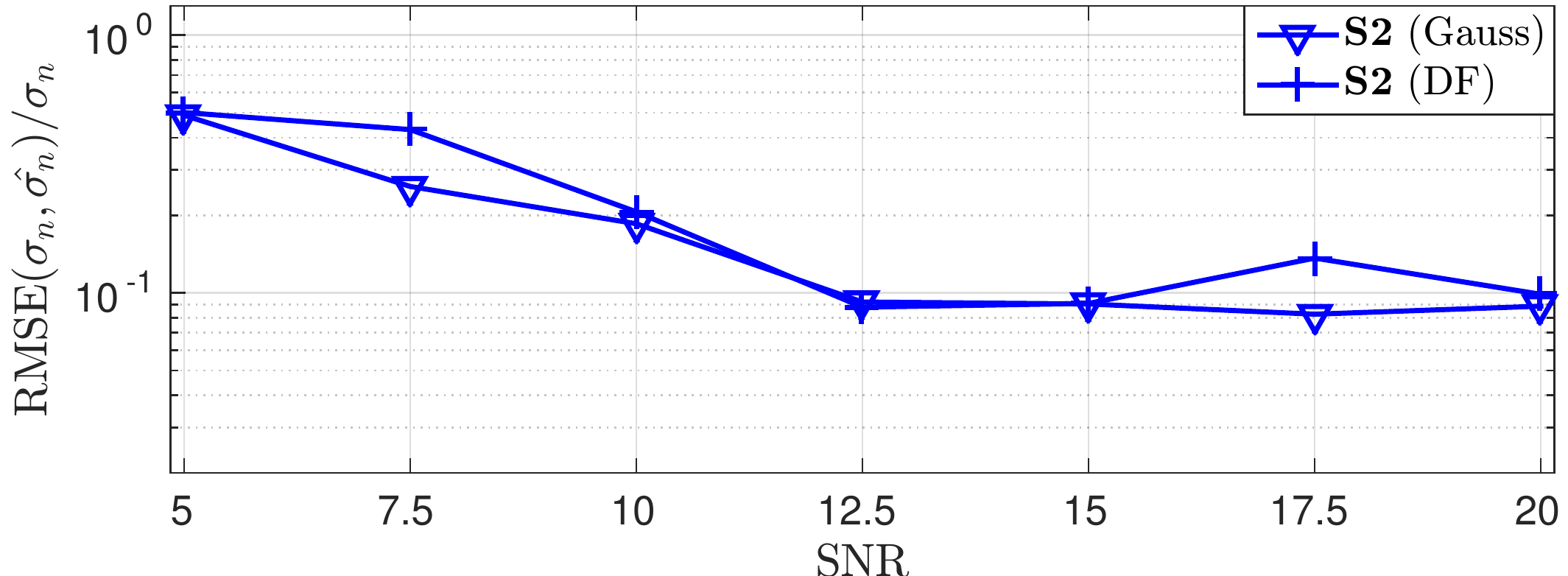}}
\caption{Performance of \textbf{\emph{S2}} in terms of the RMSE using $M/L=50\%$ of the original samples in comparison with PDL-OIAI in \cite{Weiss2016} and with the lower bound of the RMSE imposed by the CRB (RCRB).}\vspace{-0.1cm}
\label{fig:JOINT_performance_50}
\end{figure}
\begin{figure} 
\centering
    \subfigure[]{\includegraphics[width=0.75\columnwidth]{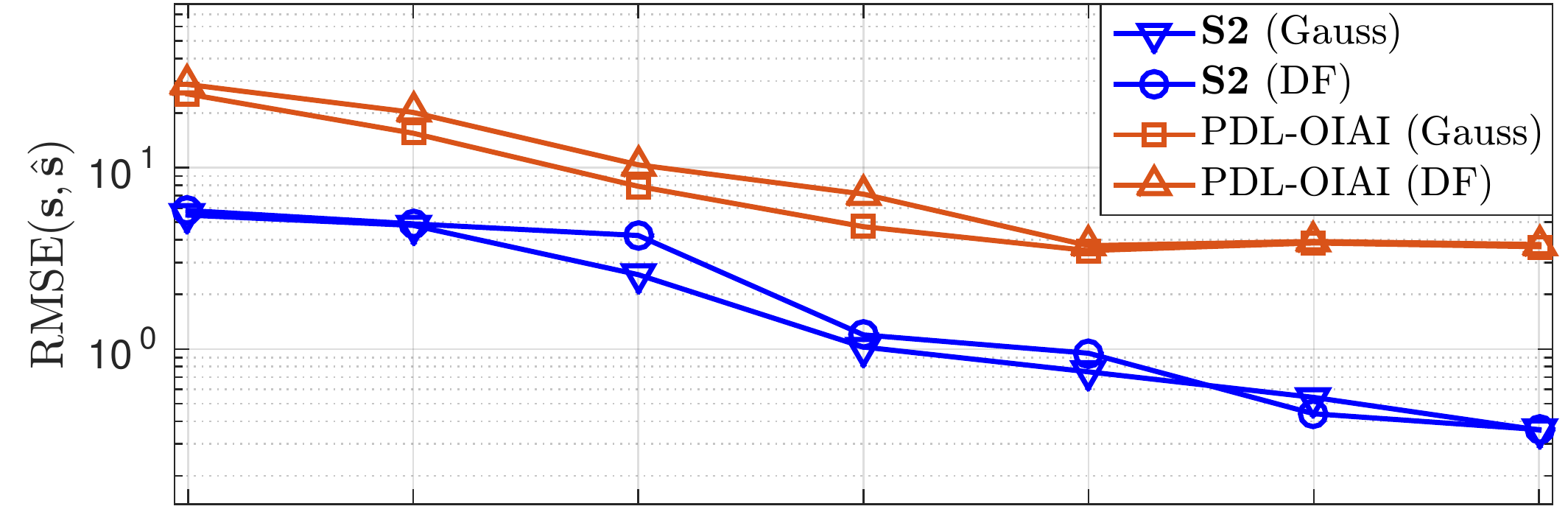}}\\[0.2cm]
    \subfigure[]{\includegraphics[width=0.75\columnwidth]{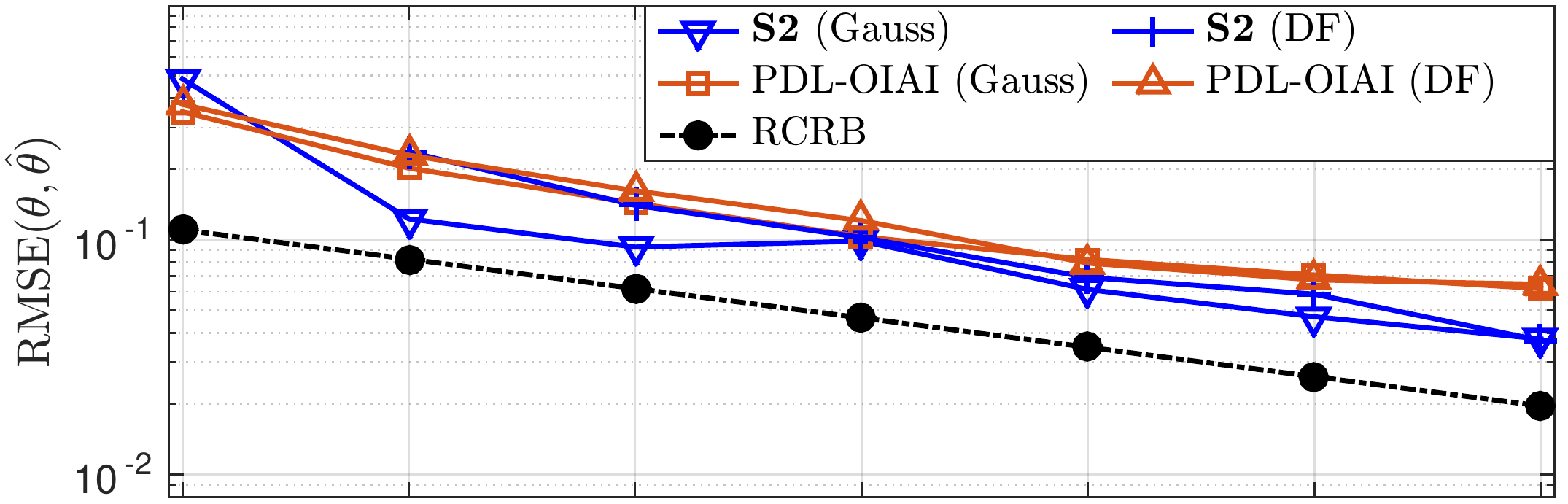}}\\[0.2cm]
    \subfigure[]{\includegraphics[width=0.75\columnwidth]{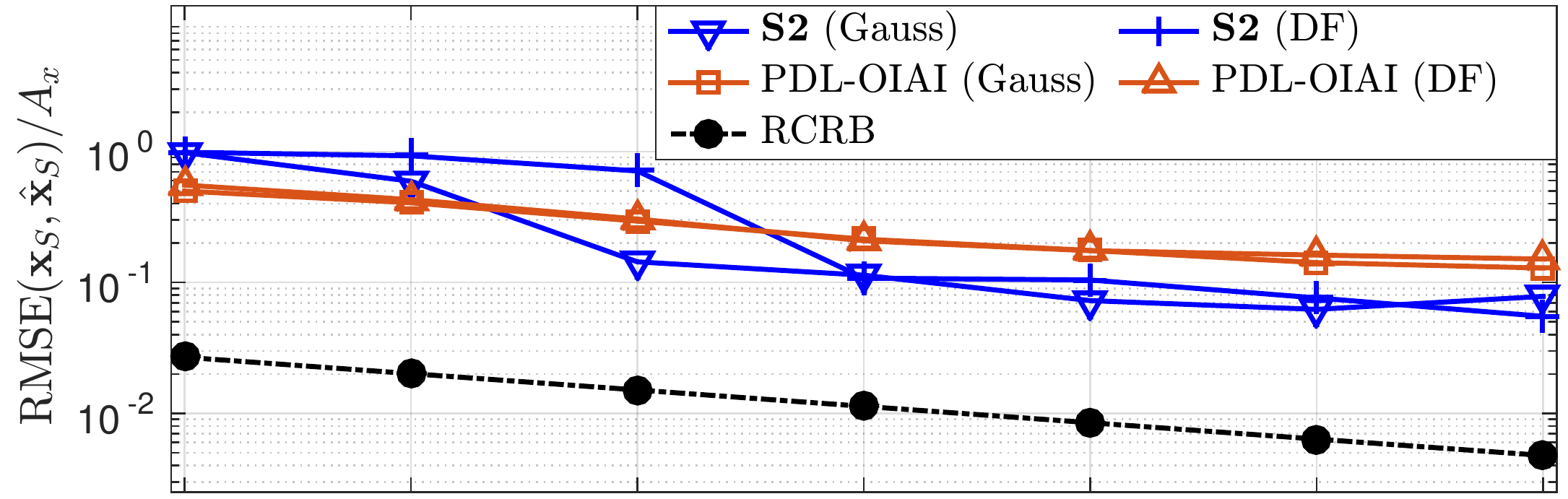}}\\[0.2cm]
    \subfigure[]{\includegraphics[width=0.75\columnwidth]{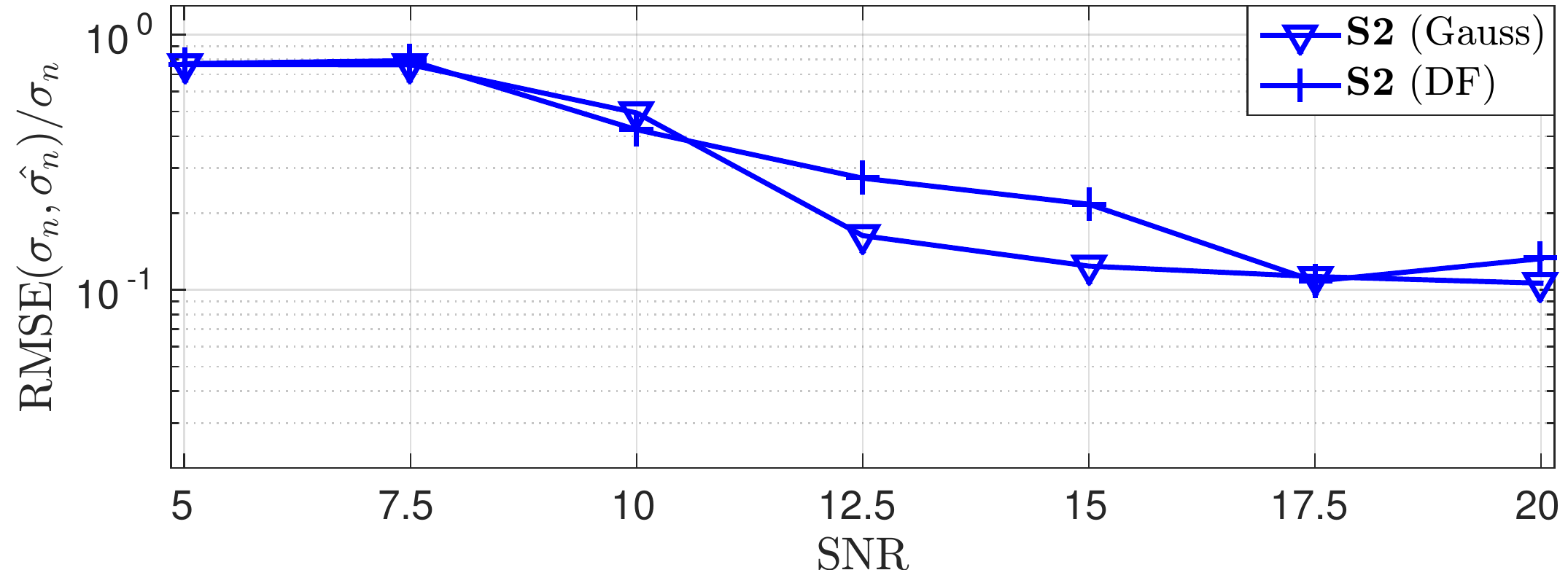}}
\caption{Performance of \textbf{\emph{S2}} in terms of the RMSE using $M/L=30\%$ of the original samples in comparison with PDL-OIAI in \cite{Weiss2016} and with the lower bound of the RMSE imposed by the CRB (RCRB).}
\label{fig:JOINT_performance_30}
\end{figure}
We define $\overline{\text{RMSE}}$ as the approximation, where the expectation is replaced by averaging estimates over 100 Monte Carlo trials. 
We compare \textbf{\emph{S1}}, \textbf{\emph{S2}} to the PDL-OIAI algorithm in \cite{Weiss2016}, which considers a deterministic 
sparse model and incorporates a pre-processing routine to handle strong dictionary coherence.
To calculate the CRB of Section \ref{sec:CRB}, the derivative of $r(t,\theta)$ with respect to $\theta$ must be determined for all
dictionary elements. Since $r(t,\theta)$ is not a simple function of $\theta$, it can be approximated for a certain value $\theta_0$. 
For the $(l,\!i)$-th element in $\mathbf{A}'(\theta)$, we obtain %
\begin{equation}
\hspace{0.01cm}	\left.\partialfrac{}{\theta}[\mathbf{a}_i(\theta)]_l\right|_{\theta_0}\!\! \approx\ \frac{r(lT_d-\tau_i,\theta_0) - r(lT_d-\tau_i,\theta_0-\Delta\theta)}{\Delta\theta}.\hspace{-0.30cm}
\end{equation}
Fig. \ref{fig:EM_performance_50}-\ref{fig:JOINT_performance_30} show the results of the proposed and the competetive method.
Herein, $\mathbf{s}\in \mathbb{N}^K$ contains all elements in $\mathcal{S}$, $\x_{\mathcal{S}}\in \mathbb{R}_+^K$ contains the coefficients of $\x$ 
with indices in $\mathcal{S}$. The $\overline{\text{RMSE}}(\x_{\mathcal{S}},\hat{\x}_{\mathcal{S}})$ compares the estimated amplitudes at the positions 
in $\hat{\mathcal{S}}$ to the true common amplitude, $A_x$, at positions in $\mathcal{S}$. 
The lower bound of the RMSE for jointly estimating \emph{deterministic} parameters $(\x_\mathcal{S},\theta)$, induced by the CRB derived in Section \ref{sec:CRB}, is denoted by 'RCRB'.\\
Fig. \ref{fig:EM_performance_50} shows the results for $\{\mathbf{s},\x_{\mathcal{S}},\theta,\sigma_n\}$, obtained by \textbf{\emph{S1}} using
$50\%$ of the original samples. For fewer samples, the EM algorithm in \textbf{\emph{S1}} becomes unstable.
\mbox{Fig. \ref{fig:JOINT_performance_50}-\ref{fig:JOINT_performance_30}} depicts the results obtained by \textbf{\emph{S2}} using $50\%$ and $30\%$
of the original samples, respectively. It shows, that \textbf{\emph{S2}} is more robust against small sample sizes and missing data than \textbf{\emph{S1}}. 
Generally, the error is only marginally affected by the type of the CS sampling matrix, i.e. (a) or (b). %
In all scenarios, \textbf{\emph{S1}} and \textbf{\emph{S2}} achieve a significantly lower error in estimating $\mathbf{s}$ than PDL-OIAI.
At low SNRs, \textbf{\emph{S1}} performs better than \textbf{\emph{S2}}, while \textbf{\emph{S2}} becomes better at high SNRs. 
However, PDL-OIAI estimates $\theta$ with slightly higher accuracy than \textbf{\emph{S1}} and \textbf{\emph{S2}}.
When $50\%$ of the original samples are used, the error closely adheres to the RCRB.  
The amplitudes, $\x_{\mathcal{S}}$, are estimated with similar accuracy by both, \textbf{\emph{S1}} and \textbf{\emph{S2}}, and no improvement 
is achieved compared to PDL-OIAI. Also, the distance to the RCRB is almost constant at all SNRs.
Regarding the noise level, $\sigma_n^2$, \textbf{\emph{S1}} yields a slightly smaller estimation error than \textbf{\emph{S2}}.
PDL-OIAI does not provide a simple means for estimating $\sigma_n^2$, which is an advantage of \textbf{\emph{S1}} and 
\textbf{\emph{S2}}. 
In the presented results for PDL-OIAI, it is assumed that pure noise samples are available to estimate $\sigma_n^2$. 
The instability of the RMSE between SNRs of 15 and 17.5 dB in Fig. \ref{fig:EM_performance_50} might arise from
averaging over an insufficient number of samples. It is also possible that the MCMC algorithm took longer to converge to the stationary distribution
for SNR$=$20 dB, e.g. due to an unlucky initialization, thus, increasing the error. %

\subsection{Experimental Data}
\begin{figure}
\centering
	\subfigure[]{\includegraphics[width=0.34\columnwidth]{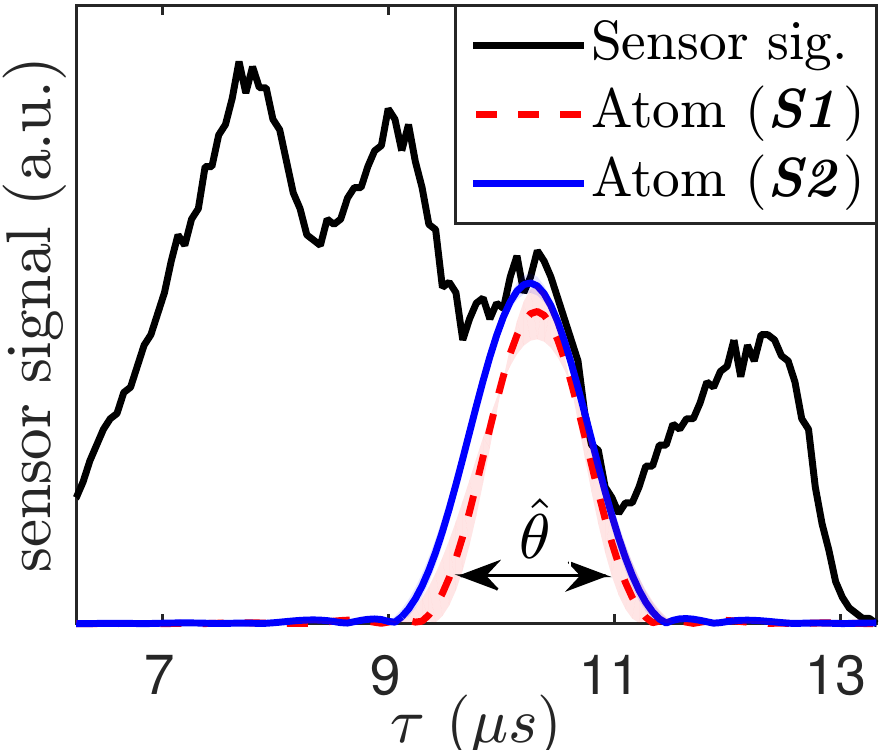}}\hspace{1.3cm}
	\subfigure[]{\includegraphics[width=0.34\columnwidth]{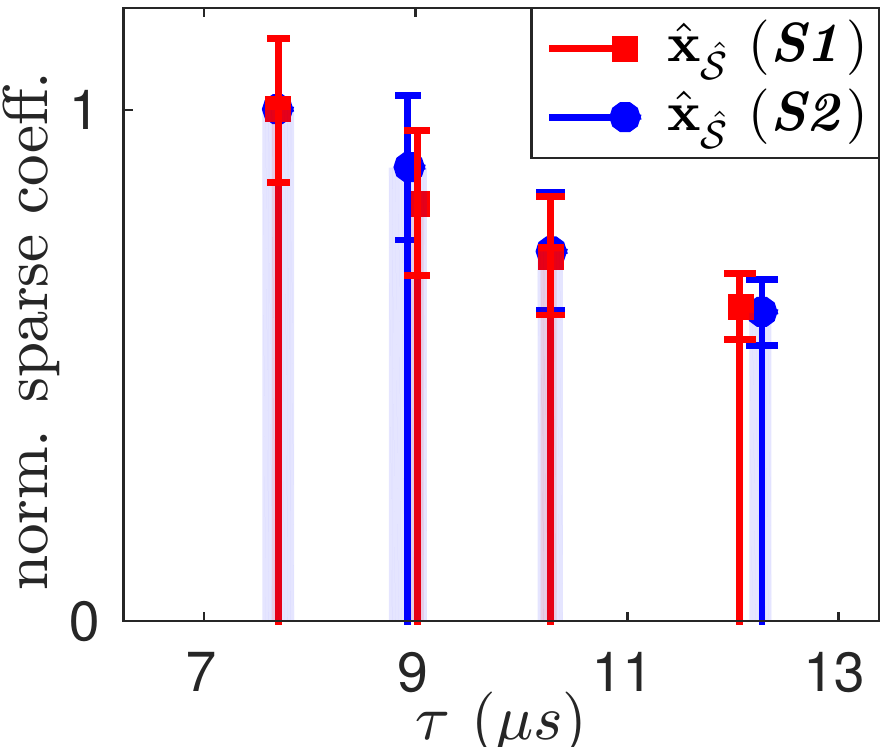}}
\caption{Real data example for \textbf{\emph{S1}} and \textbf{\emph{S2}}: Sensor signal with estimated reflections, where the shaded regions indicate the standard deviation for estimating $\theta$ (left). Estimated sparse signal, where the shaded areas show the standard error in estimating the reflection delays, indicated by $\mathcal{S}$, and the vertical error bars show the standard error for estimating $\x_{\mathcal{S}}$ (right).}
\label{fig:real_data}
\end{figure}

\noindent
To complete our study, we apply \textbf{S1} and \textbf{S2} to experimental data taken from the real fiber sensor system in \cite{Nakazaki2009,Yamashita2009}. 
It was acquired at the Yamashita laboratory of photonic communication devices at The University of Tokyo, Japan. 
We consider $L\!=\!134$ original samples of the received sensor signal and use $M/L\!=\!50\%$ of the original samples. 
The delay spacing between the $N\!\!=\!\!2L$ dictionary atoms is $\delta t\approx 50$ ns. 
The sensing fiber contains $K=4$ FBGs and the delays of the reflected signals are potentially off-grid. Their positions are 
approximately at [7.79, 9.05, 10.27, 12.30] $\mu$s. 
We perform 100 Monte Carlo trials to estimate $\{\mathcal{S}, \x_\mathcal{S}, \theta\}$.
\mbox{Fig. \ref{fig:real_data} (left)} shows the the original sensor signal and one estimated reflection from FBG$_3$. The shaded area 
indicates the standard deviation in estimating $\theta$. 
$\textbf{\emph{S1}}$ estimates a narrower reflection, which also results in slightly different estimates of $\mathcal{S}$. 
\mbox{Fig. \ref{fig:real_data} (right)} depicts $\hat{\x}_{\hat{\mathcal{S}}}$ at the estimated positions in $\hat{\mathcal{S}}$. 
The shaded areas represents the standard deviation for $\mathcal{S}$ and the 
vertical error bars indicate the standard deviation for $\x_{\mathcal{S}}$. 
Essentially, the results of \textbf{S1} and \textbf{S2} are comparable, although the variance in the estimates of $\mathcal{S}$ is 
marginally smaller in the case of $\textbf{\emph{S2}}$.
Similar performance was reported for PDL-OIAI in \cite{Weiss2016}.  

\section{Discussion}
\label{sec:discussion}
\noindent
Based on simulations and experimental results, we demonstrate that the proposed sparsity model and our DL strategies, \textbf{\emph{S1}}, \textbf{\emph{S2}}, are useful in CFS and can be used for an automated estimation of the reflection delays.
In comparison to PDL-OIAI in \cite{Weiss2016}, where the underlying model treats $\x$ and $\theta$ as deterministic parameters,
the following general observation can be made: %
The methods $\textbf{\emph{S1}}$ and $\textbf{\emph{S2}}$, based on a probabilistic sparse model, 
show comparable performance to PDL-OIAI but do not exceed the performance limit imposed by the non-Bayesian CRB. 
However, a significant improvement is achieved in estimating $\mathcal{S}$. %
It should be emphasized
\mbox{that $\mathcal{S}$} is of major importance in WDM-based CFS. 
It indicates the reflection delays, that are used to infer the quantity or nature of impairments at the FBGs.
The amplitudes, $\x_{\mathcal{S}}$, can be used to determine the sparsity level and the amount of optical power reflected from the FBGs. 
We find that all competing methods estimate $\x_{\mathcal{S}}$ similarly accurate.
The real data example shows that \textbf{\emph{S1}}, \textbf{\emph{S2}} are insensitive to 
signal features that are not explicitly modeled, e.g. the skewness of the reflections or a signal-dependent noise amplitude. 
This was also reported for \mbox{PDL-OIAI \cite{Weiss2016}.}\\ 
Our results for $\mathcal{S}$ indicate that the proposed sparsity model is better able to handle strong dictionary
coherence than PDL-OIAI, which adopts a dictionary pre-processing routine to reduce the dictionary coherence. 
We ascribe this ability in part to the favorable selective shrinkage properties of the Weibull prior. 
Such behavior was previously reported for general heavy-tailed priors in \cite{Seeger2008,Polson2010}.
Regarding the relation to non-convex optimization, we find that constraints imposed on the $\ell_p$-norm, with $0\!<\!p\!<\!1$, 
are indeed useful in the presence of strong dictionary coherence. In this context, we support the findings in \cite{Chartrand2007,Chartrand2008}, 
that report relaxed RIP conditions when $\ell_1$-minimization is replaced by non-convex optimization methods.
Another important factor, that contributes to the ability of handling strong dictionary coherence, is the local similarity model
introduced in the joint prior density of $\x$. %
We observe much sparser solutions due to its collective shrinkage property, 
without the need for any dictionary pre-processing as in PDL-OIAI.
Although this model is designed to deal with the unique features of the CFS dictionary, it can be used for general shift-invariant dictionaries with similar structures and high coherence levels. Therefore, it offers a broader applicability beyond the CFS problem.
For parametric DL, all compared methods seem equally suitable for estimating $\theta$, but \textbf{\emph{S2}} and PDL-OIAI are 
more stable for small sample sizes. 
Since the type of the CS matrix has only marginal impact, DF matrices in \cite{Achlioptas2003} are favorable. 
They are easy to implement, require low storage, and reduce the average sampling rate by 66\%, since 2/3 of all projections are zero.\\
The computational complexity of \textbf{\emph{S1}} and \textbf{\emph{S2}} is dominated by drawing samples from the posterior of $\x$ using HMC.
HMC shows high efficacy in sampling this high-dimensional space in the presence of correlation. %
It yields samples from the desired posterior, that are weakly sparse with sharp peaks only close to the true positions of the significant components. %
Compared to optimization methods such as PDL-OIAI, MCMC is slower (c.f. \cite{Mohamed2012}) but some preliminary efforts are necessary for choosing 
a proper regularization parameter in the $\ell_1$-minimization problem. %
The run-time complexity of PDL-OIAI is dominated by a costly but essential data-dependent pre-processing routine to deal with severe dictionary coherence. This can be implemented using parallel processing and might be more efficient in situations, where CFS is used for permanent perturbation 
monitoring. 
Nonetheless, it \mbox{requires an} initial estimate of the non-perturbed reference system. For this task, $\mathcal{S}$ can be more accurately 
estimated using \textbf{\emph{S1}} and \textbf{\emph{S2}}. %
In contrast to PDL-OIAI, they are also able to estimate the noise level. %
Combining these methods for calibration and permanent monitoring is a promising perspective for practical systems. %
A limiting factor in \textbf{\emph{S1}} and \textbf{\emph{S2}} is the MCMC runtime, i.e. the number of available samples for Monte Carlo
integration. Depending on the initial point, sufficient time has to be given for the algorithms to converge to the stationary distribution.  
Also, \textbf{\emph{S1}} may get stuck in local optima but
the proposed initialization using a bisectional search can lower this chance %
and helps to speeds up the convergence of the algorithm.
A possible extension of this work can include multiple CS sample vectors to improve the SNR conditions. This might yield more accurate 
results and stable behavior. A similar technique was proposed in \cite{Malioutov2005}.
Also, as pointed out in \cite{Weiss2016}, additional local dictionary parameters can be considered.
Since the reflections in the experimental data are non-uniform, this might improve both robustness and accuracy.

\section{Conclusion}
\label{sec:conclusion}
\noindent
We present a sparse estimation and parametric dictionary learning framework for Compressed Fiber Sensing (CFS) based on a probabilistic 
hierarchical sparse model. 
The significant components in the sparse signal indicate reflection delays, that can be used to 
infer the quantity and nature of external impairments. %
In order to handle severe dictionary coherence and to accomodate specific characteristics of the signal, %
a Weibull prior is employed to promote selective shrinkage. 
This choice can be related to non-convex optimization based on the \mbox{$\ell_p$-norm.}
To further alleviate the problem of dictionary coherence, we leverage the particular structure of the dictionary and assign a 
local variance to the differential sparse coefficients.
This model can be useful for general shift-invariant dictionaries with similar structure and strong coherence.
We propose two parametric dictionary learning strategies, \textbf{\emph{S1}} and \textbf{\emph{S2}}, to estimate the dictionary parameter, $\theta$. 
In $\textbf{\emph{S1}}$, $\theta$ is treated as a deterministic parameter and estimated using a Monte Carlo EM algorithm. 
In $\textbf{\emph{S2}}$, a probabilistic hierarchical model for $\theta$ is considered. %
A hybrid MCMC method based on Gibbs sampling and Hamilton Monte Carlo is used for approximate inference. 
In simulations and by experimental data, we show the applicability and efficacy of the proposed sparse model, together with 
the methods \textbf{\emph{S1}} and \textbf{\emph{S2}}, for an automated estimation of the reflection delays and the dictionary parameter in CFS.
In a comparative analysis with an existing method, based on a deterministic sparse model, we highlight advantages, disadvantages and 
limitations, that can serve as a guidance to choose an adequate method for practical systems. 
To better assess the performance gain of a probabilistic sparse model, the Cram{\'e}r-Rao bound is derived for the joint estimation of deterministic 
sparse coefficients and the dictionary parameter in CFS. %
Drawbacks of the proposed methods are the generally high computational costs of MCMC methods, and the lack of simple diagnostic tools for Markov chain convergence and sample independence. Also, \textbf{\emph{S1}} suffers from the problem of local optima. %
As a remedy, we propose a bisectional search to find a proper initialization. 
In subsequent investigations, multiple CS sample vectors and additional local dictionary parameters can be taken into account. 
Also, variational Bayes methods can be used to speed up computations. %

\section*{Acknowledgment}
\noindent
This work was supported by the\! 'Excellence Initiative'\! of the German
Federal and State Governments and the Graduate School
of Computational Engineering at Technische Universit{\"a}t Darmstadt.\\
\noindent
The authors would like to thank Professor S. Yamashita and his
group at The University of Tokyo, Japan, for kindly providing
experimental data of the fiber sensor \mbox{in \cite{Yamashita2009}.}

 \bibliographystyle{plain}
\bibliography{./bibliography}

\end{document}